\documentclass[sigconf,screen]{acmart}
\usepackage[english]{babel}
\usepackage{blindtext}
\usepackage{amssymb} 
\usepackage{fontspec}
\usepackage{multirow}
\usepackage{fvextra}
\usepackage{subcaption}
\usepackage{enumitem}
\usepackage[dvipsnames]{xcolor}
\usepackage{makecell}
\usepackage{pifont}
\usepackage{array}
\usepackage{listings}   
\usepackage{upquote}        
\usepackage{tcolorbox}
\tcbuselibrary{skins,breakable}
\usepackage{fancyvrb}
\usepackage{tikz}
\usetikzlibrary{positioning,calc}
\usepackage{algpseudocode}
\usepackage{cleveref}
\usepackage{quoting}

\makeatletter
\renewcommand{\@date}{} 
\makeatother
\makeatletter
\renewcommand{\sectionautorefname}{\S\@gobble}
\makeatother 
\makeatletter
\renewcommand{\subsectionautorefname}{\S\@gobble}
\makeatother 
\makeatletter
\renewcommand{\subsubsectionautorefname}{\S\@gobble} 
\makeatother

\AtBeginDocument{%
  \providecommand\BibTeX{{%
    \normalfont B\kern-0.5em{\scshape i\kern-0.25em b}\kern-0.8em\TeX}}%
}
\makeatletter
\renewcommand{\sectionautorefname}{\S\@gobble}
\makeatother 
\makeatletter
\renewcommand{\subsectionautorefname}{\S\@gobble}
\makeatother 
\makeatletter
\renewcommand{\subsubsectionautorefname}{\S\@gobble} 
\makeatother

\lstdefinestyle{pythonstyle}{
  language=Python,
  basicstyle=\ttfamily\scriptsize,
  keywordstyle=\color{blue},
  commentstyle=\color{teal},
  stringstyle=\color{brown},
  showstringspaces=false,
  breaklines=true,
  frame=single,
  numbers=none,
  tabsize=2,
  columns=fullflexible
}
\DeclareCaptionType{listing}[List of Listings][Listing] 
\usepackage{cuted}

\newcommand{\TheSystem}{Engram\xspace} 
\newcommand{\name}{\TheSystem}
\newcommand{\archive}{Archive\xspace}
\newcommand{\Ledger}{Research Digest\xspace}
\newcommand{\ledger}{research digest\xspace}

\newcommand{\Fig}[1]{Fig.~\ref{fig:#1}}
\newcommand{\NewPara}[1]{\par\medskip
\textnormal{\bf#1}\nobreak}

\newcommand{\eat}[1]{}

\setlist[itemize]{nosep,left=1pt}
\setlist[enumerate]{noitemsep, topsep=2pt, parsep=0pt, partopsep=0pt, leftmargin=*}

\lstdefinestyle{commandstyle}{
    language=Python,
    backgroundcolor=\color{gray!10},
    keywordstyle=\color{blue},
    stringstyle=\color{purple},
    commentstyle=\color{green!50!black},
    basicstyle=\ttfamily\scriptsize,
    breaklines=true,
    frame=single,
    framerule=0pt,
    framesep=5pt,
    rulecolor=\color{black},
    numbersep=5pt
}

\lstdefinestyle{bashstyle}{
    language=bash,
    backgroundcolor=\color{gray!10},
    keywordstyle=\color{blue},
    stringstyle=\color{purple},
    commentstyle=\color{green!50!black},
    basicstyle=\ttfamily\scriptsize,
    breaklines=true,
    frame=single,
    framerule=0pt,
    framesep=5pt,
    rulecolor=\color{black},
    numbersep=5pt
}

\lstdefinestyle{outputstyle}{
    backgroundcolor=\color{blue!5},
    basicstyle=\ttfamily\scriptsize,
    frame=tblr, 
    framesep=8pt,
    breaklines=true,
}

\lstdefinestyle{conversationstyle}{
    basicstyle=\sffamily\small,   
    backgroundcolor=\color{green!10},
    breaklines=true,
    frame=tblr,                   
    framerule=1pt,                
    rulecolor=\color{blue!60},    
    xleftmargin=3pt,             
    framexleftmargin=10pt,        
    breakindent=0pt,
}

\usepackage{pgfplotstable}
\usepackage{tikz}
\usetikzlibrary{fit,backgrounds,decorations.pathreplacing}
\usepgfplotslibrary{groupplots,statistics,fillbetween}
\pgfplotsset{compat=1.18}

\usetikzlibrary{
  shapes.misc, shapes.geometric, shapes.multipart, shapes.arrows,
  positioning, quotes, arrows.meta, decorations.pathmorphing, 
  matrix, graphs, fit, calc, automata, shadows
}

\DeclareGraphicsExtensions{.pdf,.png,.jpg}
\usepackage{epstopdf}
\usepackage{varwidth}

\usepackage{caption}
\usepackage[authormarkup=none,commandnameprefix=ifneeded]{changes}

\acmSubmissionID{}

\renewcommand\shortauthors{}
\setcopyright{none}
\renewcommand{\footnotetextcopyrightpermission}[1]{}

\begin{document}
\title{Improving Coherence and Persistence in Agentic AI \\ for System Optimization}

\settopmatter{authorsperrow=4}

\author{Pantea Karimi}
\authornote{Equal contribution}
\affiliation{
  \institution{MIT}
  \city{Cambridge}
  \country{USA}
}

\author{Kimia Noorbakhsh}
\authornotemark[1]
\affiliation{
  \institution{MIT}
  \city{Cambridge}
  \country{USA}
}

\author{Mohammad Alizadeh}
\affiliation{
  \institution{MIT}
  \city{Cambridge}
  \country{USA}
}

\author{Hari Balakrishnan}
\affiliation{
  \institution{MIT}
  \city{Cambridge}
  \country{USA}
}

\renewcommand{\shortauthors}{}

\begin{sloppypar}
    
\begin{abstract}
    Designing high-performance system heuristics is a creative, iterative process requiring experts to form hypotheses and execute multi-step conceptual shifts. While Large Language Models (LLMs) show promise in automating this loop, they struggle with complex system problems due to two critical failure modes: {\em evolutionary neighborhood bias} and the {\em coherence ceiling}. Evolutionary methods often remain trapped in local optima by relying on scalar benchmark scores, failing when coordinated multi-step changes are required. Conversely, existing agentic frameworks suffer from context degradation over long horizons or fail to accumulate knowledge across independent runs. 

We present \textbf{\name}, an agentic researcher architecture that addresses these limitations by decoupling long-horizon exploration from the constraints of a single context window. \name organizes exploration into a sequence of agents that iteratively design, test, and analyze mechanisms. At the conclusion of each run, an agent stores code snapshots, logs, and results in a persistent {\em \archive} and distills high-level modeling insights into a compact, persistent {\em \Ledger}. Subsequent agents then begin with a fresh context window, reading the \Ledger to build on prior discoveries. 

We find that \name exhibits superior performance across diverse domains including multi-cloud multicast, LLM inference request routing, and optimizing KV cache reuse in databases with natural language queries.
\end{abstract}

\maketitle
\pagestyle{plain}

\section{Introduction}
\label{s:intro}

Designing high-performance heuristics and algorithms for computer systems is a creative and iterative process. Experts form hypotheses about system bottlenecks, implement candidate mechanisms, test them under realistic workloads, and use the findings to refine the design, often through multi-step conceptual shifts rather than code tweaks. Recent work has explored using large language models (LLMs) to automate this loop~\cite{glia,cheng2025let}, yet prior approaches struggle to reliably produce expert-level solutions to complex systems problems.

We identify two critical failure modes:
\begin{enumerate}
    \item \emph{Evolutionary neighborhood bias}: Code-evolution systems propose code variants and select them using a scalar benchmark score~\cite{FunSearch,AlphaEvolve,openevolve,lange2025shinkaevolve,EoH,agrawal2026gepa}. This approach can work when progress comes from incremental refinements to a stable template, but it often fails when improvements require coordinated multi-step changes; e.g., reformulating the problem, adding tractable relaxations, or accepting temporary regressions while moving to a different algorithmic family.
    
    \item The \emph{coherence ceiling}: While agentic frameworks such as Glia~\cite{glia} enable hypothesis formation and targeted experimentation, they struggle with long-horizon design. A single long-running context eventually suffers from degradation and ``context rot,'' where attention becomes uneven~\cite{hong2025context}. Conversely, independent ``best-of-$N$'' runs do not accumulate knowledge; each run must rediscover identical modeling insights from scratch. Code evolution methods, on the other hand, lack long-horizon coherence because each LLM invocation is unaware of the {\em thought process} behind previous attempts. 
    
\end{enumerate}

We present \textbf{\name}, an agentic researcher architecture that overcomes these limitations by decoupling long-horizon exploration from the constraints of a single context window. \name organizes exploration into a sequence of agents that iteratively design, test, and analyze mechanisms. At the conclusion of each run, an agent archives code snapshots, execution logs, and experimental results into a persistent \textit{\archive}. Crucially, \name introduces a \textit{structured handoff} wherein each agent distills high-level insights, findings, and failure diagnoses into a compact \textit{\Ledger}. Every subsequent agent begins with a fresh context window, reading the \Ledger to build upon prior discoveries and findings. This architecture enables \name to sustain a coherent research exploration across hundreds of trials, bypassing the performance degradation typical of single-context agentic threads.

\begin{figure*}[t]
    \centering
    \includegraphics[width=0.9\linewidth]{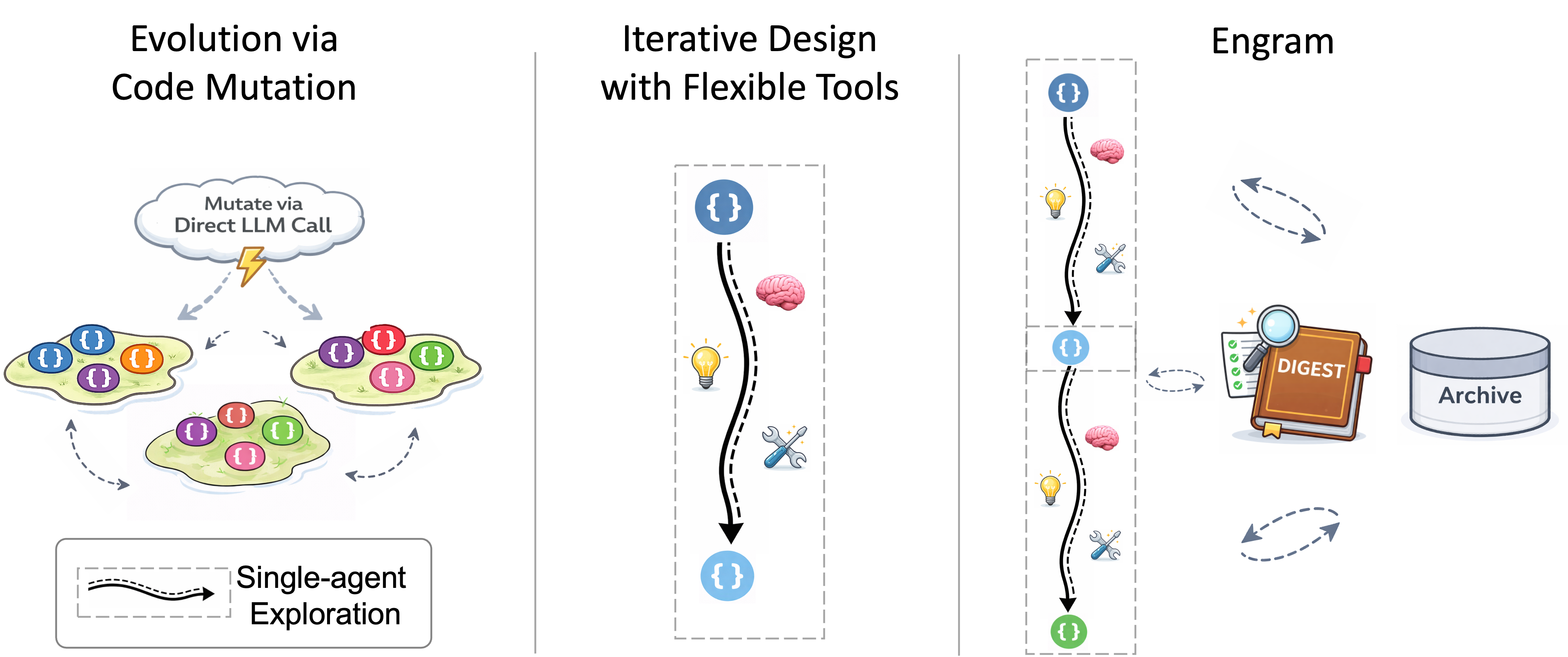}
    \caption{The three paradigms for LLM-based heuristic design. Evolutionary approaches with code mutation invoke an LLM with a predefined context format, mutating and selecting candidates based on scalar scores. Iterative design with flexible tool access (e.g., Glia) performs coherent experiment-guided exploration, but each exploration is restricted to a bounded LLM context window.  \name combines agent explorations with a shared \ledger that persists insights across explorations (\autoref{sec:design}), imporving persistence while preserving long-horizon coherence and flexibility.}
    \label{fig:method_threads}    
\end{figure*}

Our contributions are:
\begin{itemize}

   \item \textbf{Enhanced long-horizon coherence.} We introduce a structured handoff and archival mechanism that enables cumulative progress across agents. The persistent \Ledger ensures that modeling insights and failure diagnoses persist beyond the lifetime of any single agent exploration.
   \item \textbf{Discovery of new system heuristics that surpass prior state-of-the-art.} \name discovers novel heuristics across multiple domains. 
In multi-cloud multicast~\cite{cloudcast}, \name synthesizes new multicast heuristics that achieve a best overall cost of \$622, surpassing the reported human state-of-the-art (SOTA) of \$626 and all evolutionary baselines. 
For LLM inference request routing~\cite{vidur_base_global_scheduler}, \name discovers improved strategies that reduce mean response time to 23.9\,s, outperforming the expert-designed heuristic as well as Glia~\cite{glia} (25.7\,s) and all evolutionary baselines.

\item \textbf{Outperforming prior approaches across diverse problems.} We evaluate \name on nine systems problems -- eight from ADRS benchmark~\cite{UCB_ADRS_GitHub} and LLM request router~\cite{glia}. \name outperforms human SOTA in eight of the nine settings and exceeds (7 of 9) or matches (2 of 9) OpenEvolve in all evaluated categories.

\end{itemize}
\section{Why LLMs Struggle on System Optimization Problems}
\label{sec:background}

\begin{table}[t]
\centering
\small
\setlength{\tabcolsep}{4pt}
\begin{tabular}{@{}l ccc@{}}
\toprule
\textbf{Approach} & \textbf{Coherence} & \textbf{Flexibility} & \textbf{Persistence} \\
\midrule
Evolutionary code mutation & Low & Low & High \\
Iterative design via flexible tools & High & High & Low \\
\textbf{\name} & \textbf{High} & \textbf{High} & \textbf{High} \\
\bottomrule
\end{tabular}
\caption{Trade-offs among LLM-based heuristic design. Code-mutation evolution is persistent is limited is coherence and flexibility; tool-based agents enable coherent and flexible exploration but have limited persistence. \name achieves all three.}
\vspace{-20 pt}
\label{tab:comparision}
\end{table}

Optimizing systems and designing heuristics is fundamentally a creative, multi-step process. High-performance mechanisms rarely emerge from local code tweaks; instead, they require formulating hypotheses about bottlenecks, constructing and conducting experiments, interpreting experimental results, and executing multi-step conceptual shifts~\cite{glia}.

Recent research has sought to use LLM-based agents to automate this loop~\cite{glia, cheng2025let}. We categorize current approaches for such heuristic design into two paradigms and analyze why they struggle to sustain long-horizon progress. These paradigms are:
(i) evolution via code mutation (e.g., OpenEvolve~\cite{openevolve}) and (ii) iterative design with reasoning and tools 
(e.g., Glia~\cite{glia}). We analyze both along three criteria:

\begin{itemize}
\item \textbf{Context coherence:} Are agent decisions informed by relevant findings and the thought process behind previous attempts? 
\item \textbf{Flexibility:} Can the agent take free-form actions (e.g., run code, inspect data, use tools) instead of a fixed-prompt format? 
\item \textbf{Persistence:} Can the search continue for long horizons without quality degrading due to context growth?
\end{itemize}

Each paradigm satisfies some, but not all, of these criteria  (\autoref{tab:comparision}). 

\paragraph{Evolution via code mutation.} In this paradigm, an LLM mutates code candidates that are evaluated against a benchmark to produce a scalar score, which in turn guides selection for subsequent generations. FunSearch~\cite{FunSearch}, AlphaEvolve~\cite{AlphaEvolve}, OpenEvolve~\cite{openevolve}, Evolution of Heuristics~\cite{EoH}, and GEPA~\cite{agrawal2025gepa} follow this pattern, with proposals like ADSR~\cite{cheng2025let} showing how systems problems can be tackled using frameworks such as OpenEvolve. These approaches query the LLM with a fixed prompt template built from prior candidates and scores. High-scoring codes survive to seed the next generation. 

System optimization is a deliberative process requiring sustained reflection and the ability to learn from intermediate failures. Fixed templates providing only snapshots of prior code and scores cannot encode a designer's evolving line of reasoning. For instance, a designer may conduct experiments on a suboptimal intermediate solution to gather diagnostic data; however, code evolution methods often prune such candidates because their immediate scores worsened, and subsequent LLM calls lack context regarding the underlying thought process behind such experiments.

\begin{figure*}[t]
  \centering
  \includegraphics[width=\linewidth]{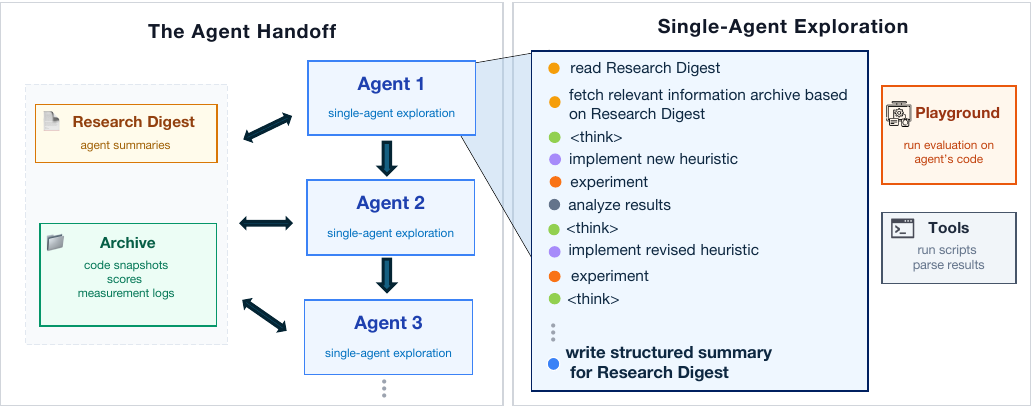}
  \caption{\name's design is based on a sequence of reasoning-based agent explorations that produce and evaluate ideas based on hypotheses driven from experimental data analysis. Each agent begins by analyzing the problem and reviewing the \ledger summarizing the findings of the previous agents, using that information to formulate its own exploration and experimentation plan. The agent executes this plan through design, experimentation, and analysis. Upon completion, it writes a summary of its findings to the \ledger, stores all the details in the Archive, and hands of the research process to the next agent. The process typically terminates when the research budget is exhausted.}
  \label{fig: workflow}
\end{figure*}

\paragraph{Iterative design with reasoning and flexible tools.}
Inspired by coding agents such as Codex \cite{openai_codex_product},
 Glia~\cite{glia} takes a alternate path to help with flexibility and coherence.  
Each agent exploration\footnote{An agent exploration is defined as the sequence of actions (e.g., read, write, and tool use) performed by an LLM-based agent from the beginning of its exploration until termination.} works in a programming environment with tool access, exploring the design problem through a coherent sequence of actions (see~\autoref{fig:method_threads}). Unlike evolutionary methods, the agent can run experiments on a testbed or simulator, execute shell commands, analyze and reason over experimental data, and iteratively refine designs. A weakness of this approach is that context growth will eventually hit a limit or degrade the quality of agent performance in long-range thinking.
Glia~\cite{glia} proposes launching multiple independent agents sequentially or in parallel to mitigate this weakness. However, these agents do not share any knowledge among themselves, each agent often rediscovers prior insights, limiting long-range progress.

Comparing these two approaches, the evolutionary approach has persistence but has weak coherence and flexibility, whereas the second approach is flexible and coherent but not persistent (see \autoref{tab:comparision}). We see these limitations in our analysis of a cloud multicast scheduling problem in \autoref{sec:cloudcast}.

\name bridges this gap with a shared \Ledger that preserves insights across explorations, achieving coherence, flexibility, and persistence. \autoref{fig:method_threads} illustrates the three approaches.

\section{\name Design}
\label{sec:design}

\name combines two key ideas:
\begin{enumerate}
    \item explore using the scientific method to develop ideas (\textsc{hypothesize $\rightarrow$ implement $\rightarrow$ experiment $\rightarrow$ analyze $\rightarrow$ hypothesize} cycle)~\cite{glia} and
    
    \item create compact, structured knowledge and transfer it to subsequent agents to achieve the long-horizon coherence needed for successful ideation. 
    
\end{enumerate}

The \name structure (\autoref{fig: workflow}) is a sequence of LLM agents that each work on the heuristic design problem using reasoning methods (\autoref{sec:coherent-agent}), with each agent \emph{handing off} its findings to the next via the \ledger (\autoref{sec:handoff}). The coherent exploration step aims to ensure that agents understand why their designs succeed or fail, rather than merely generating candidate solutions.  
The structured knowledge creation and transfer step ensures that this understanding persists beyond each agent’s lifetime and is passed to its successor, enabling knowledge to accumulate without any single agent suffering from context degradation.

\begin{figure}[t]
  \centering
  \includegraphics[width=\linewidth]{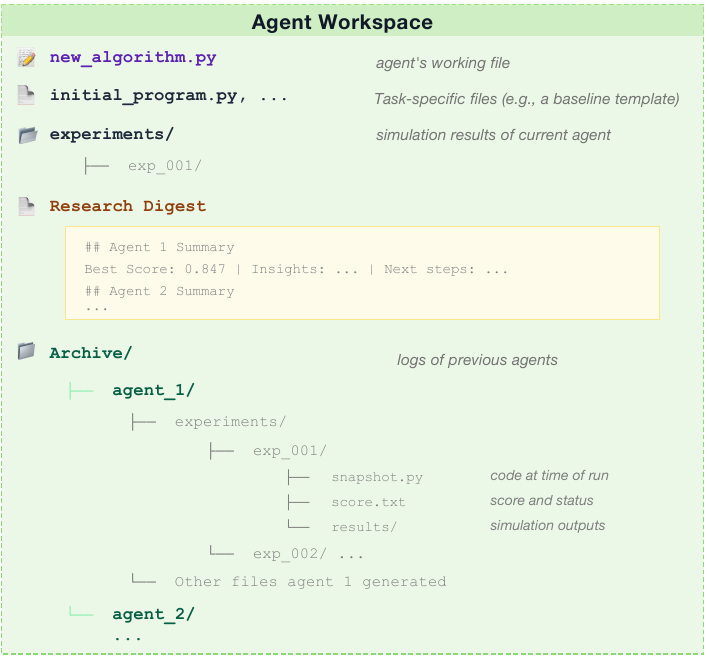}
  \caption{Workspace of a single agent in \name.}
  \label{fig: agent-workspace}
\end{figure}

\subsection{Single Agent Exploration}
\label{sec:coherent-agent}

Each agent operates within a workspace containing task-specific artifacts provided by the user (e.g., documents, sample workloads, baseline implementations, etc.). The agent may create and modify files within this workspace while developing its solution. Additionally, the agent has access to a tool that accepts code files as input and runs it within an evaluation playground (e.g., a simulator or an experimental testbed), and returns the resulting outputs. The agent can run shell commands, read, write, and edit files, and search through the contents of its workspace (in the future, it can be extended to perform external research as well). The playground outputs include log files and performance metrics that the agent can analyze to compute statistics and diagnose performance issues. In addition, crucially, each agent can access the workspaces and research digests produced by previous agents (\autoref{sec:handoff}). Importantly, the \ledger and \archive are stored as external artifacts rather than being embedded directly in the LLM’s context window. Agents retrieve relevant portions of these artifacts on demand through tool calls, ensuring that prior history does not consume context tokens unless explicitly accessed.

Concretely, each agent follows a structured research agenda. Through the system prompt (\autoref{fig:agent_lifecycle_prompt}), it is instructed to begin by reviewing the problem specification and any available prior work, then articulating a plan and a specific hypothesis describing what approach to try and why. The agent implements the idea in the working code file, runs a simulation, and performs a structured post-experiment analysis: comparing the new results to the prior ones, verifying that the implementation executed correctly, and identifying sources of improvement or limitations to guide further refinement. The \textsc{hypothesize $\rightarrow$ implement $\rightarrow$ experiment $\rightarrow$ analyze} cycle continues until the agent decides that it is ready to conclude its exploration~\cite{glia}. 

We implement this observe-before-changing discipline through the system prompt which mirrors the practice of a careful experimental researcher. Once finished, the agent appends a summary to the \ledger that serves as guidance for subsequent agents, documenting the attempted approaches, key insights gained, recommended next steps, and strategies that proved ineffective.

\subsection{The Agent Handoff}
\label{sec:handoff}

On a handoff, \name instantiates a new agent with an empty context window and provides it with the task description and access to the \ledger and the \archive. The agent conducts the exploration process described in~\autoref{sec:coherent-agent}. At the end, the agent writes all the details into the \archive including code snapshots, logs, and results. It also writes a structured summary of findings to the persistent \ledger. ~\autoref{fig: agent-workspace} depicts the workspace presented to a newly instantiated agent.

The handoff is a deliberate architectural choice that refreshes LLM context. A single long-running agent accumulates numerous tokens over the course of its exploration. As this context grows, the model’s attention becomes increasingly uneven across earlier content, a well-documented limitation of current LLMs \cite{liu-etal-2024-lost, hong2025context, du-etal-2025-context}. Handoff overcomes this problem without losing key information: when an agent has exhausted its productive capacity, it distills what it has learned into external artifacts, and a fresh agent inherits those artifacts with a clean context window.

\name's design balances two competing strengths. A single, continuous context provides rich coherence and flexibility that prevents long-context degradation, while handoff introduces structured context management that achieves long-horizon persistence. Each successor agent begins with a compact, focused statement of the problem and prior findings rather than a long and deteriorating interaction history. 

We have implemented \name using the \textsc{deepagents}\footnote{\url{https://github.com/langchain-ai/deepagents}} library built on LangChain and LangGraph \cite{chase2022langchain}. 
\section{Case Studies}
\label{sec:Methodology}
\begin{table}[t]
\centering
\resizebox{\columnwidth}{!}{
\begin{tabular}{lll}
\toprule
\textbf{System} & \textbf{Parameter Name} & \textbf{Value} \\
\midrule
\multirow{1}{*}{EoH} 
  & Initial population size (N) & 20 \\
\midrule
\multirow{1}{*}{FunSearch} 
  & \# Prompt programs (k) & 3 \\
\midrule
\multirow{4}{*}{OpenEvolve (Vidur)} 
  & Number of islands & 6 \\
  & Exploration probability & 0.6 \\
  & Exploitation probability & 0.4 \\
  & Initial program & LLQ router \\
\midrule
OpenEvolve (Others) 
  & Configuration & Benchmark settings~\cite{cheng2025barbariansgateaiupending} \\
\midrule
Glia 
  & \# Parallel threads & 4 \\
\bottomrule
\end{tabular}
}
\caption{Experimental settings and hyperparameters.}
\label{tab:params}
\end{table}

We study \name on three diverse system problems to show generality across optimization structure, codebases, and performance objectives. Along with the primary metric for each problem, we report the 90\% bootstrapped confidence interval. We run each approach 10 times and each run has a budget of 100 evaluation runs. 

We compare \name to four state-of-the-art frameworks that use LLMs for discovery: (i) Evolution of Heuristics (EoH)~\cite{EoH}, (ii) FunSearch~\cite{FunSearch}, (iii) OpenEvolve~\cite{openevolve}, and (iv) Glia~\cite{glia}.~\autoref{tab:params} shows the parameters for these frameworks.

Evolution of Heuristics (EoH)~\cite{EoH} incorporates a ``thought'' phase into algorithm design and iteratively refines candidate heuristics using five operators: crossover operators that generate diversity and recombine ideas, and mutation operators that improve, tune parameters, or simplify heuristics.

FunSearch~\cite{FunSearch} uses a best-shot optimization paradigm within an island model. Each generation refines top-performing programs selected from the solution database, maintaining a bounded candidate pool. We limit each island to 20 algorithms.

OpenEvolve~\cite{openevolve}, an open-source implementation of AlphaEvolve~\cite{AlphaEvolve}, adopts an island-based evolutionary framework. Beginning with a seed program and a task-specific scoring function, it uses a prompt sampler to generate LLM inputs that produce program variants. Evolutionary selection and migration across islands drive improvement over successive generations.

Glia~\cite{glia} is an agentic framework that uses reasoning-driven experimentation to produce candidates: the agent forms hypotheses, runs targeted experiments, analyzes outcomes, and iterates on the candidate algorithm. In our evaluation, we use the multi-context variant of Glia that takes the best of $N$ independent runs at test time ($N=4$).

\subsection{Case Study: Multi-Cloud Multicast}
\label{sec:cloudcast}

\begin{figure}[t]
    \centering
    \includegraphics[width=\linewidth]{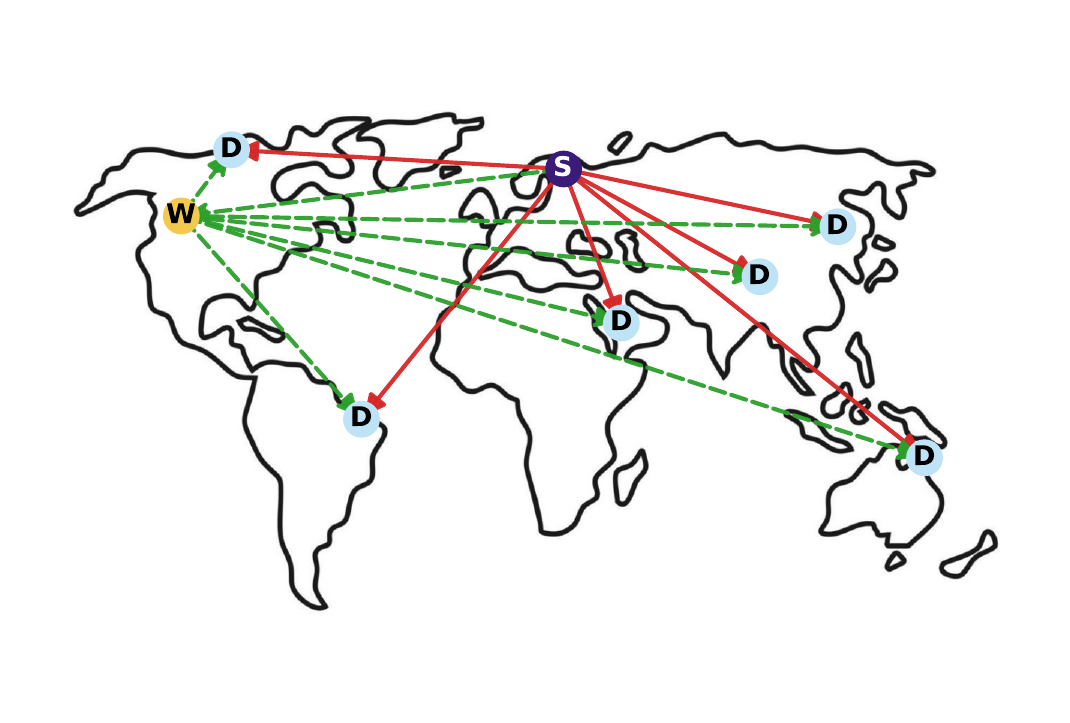}
    \vspace{-30pt}
    \caption{Multi-cloud data replication from a source (purple) to destinations (blue) across the world, either directly or via a waypoint (yellow) to avoid expensive or slow links.\protect\footnotemark}
    \label{fig:multicast}
\end{figure}
\footnotetext{Image reproduced by the authors inspired from Cloudcast~\cite{cloudcast}.}

\NewPara{Problem overview.}

Modern cloud systems replicate large datasets across geographically distributed regions for geo-replicated storage, analytics, machine learning model distribution, and disaster recovery~\cite{280908,280902,wu2013spanstore}. As shown in \autoref{fig:multicast}, data may be sent directly from a source to each destination or routed through intermediate waypoints. Waypoints can reduce end-to-end completion time and monetary cost by avoiding expensive egress links.

At first glance, this resembles classical multicast. The cloud setting, however, changes the design space~\cite{marks2015high}. Transfers are governed by asymmetric and policy-driven egress pricing rather than purely technical constraints~\cite{cloudcast}. Moreover, clouds permit elastic provisioning: one can create transient waypoints, which introduces a placement decision coupled with routing. Our goal is to design a delivery plan that minimizes the cost of sending the data under a time budget.

\NewPara{Human SOTA.}
Prior work~\cite{cloudcast} has formulated this problem as an optimization (see \autoref{appendix:cloudcast_full_optimization}). 
However, solving this optimization problem directly is impractical at even modest scale. For example, in the 71-node setting in our experiments, there are $2^{50410}$ possible configurations; a practical solution must therefore rely on heuristics and approximations~\cite{cloudcast}. Designing effective heuristics for such optimization problems is not trivial~\cite{cloudcast}; in fact, even formulating this problem as in \autoref{appendix:cloudcast_full_optimization} is a creative process. Cloudcast~\cite{cloudcast} is the human state-of-the-art solution to this problem. It uses domain-specific reductions to preserve the optimization structure while making it tractable.

\NewPara{Benchmark.} We use the same benchmark as the Cloudcast in ADRS~\cite{cheng2025barbariansgateaiupending}. The benchmark uses a 71-node directed topology, with each edge annotated by egress price and measured throughput, and evaluates five broadcast configurations with fixed source/destination sets under provider-specific ingress/egress limits and per-region VM caps. We have implemented a verifier to check candidate heuristics for correctness and then score them by total cost (egress + VM cost), under a fixed time budget.

\NewPara{AI task formulation.}
Our goal is to evaluate whether an agent can \emph{discover} effective algorithms for multi-cloud multicast. To understand how much guidance is needed, we provide the agent with three different task descriptions at different levels of detail: 

\begin{itemize}
\item \textbf{Minimal prompt:} a generic ``write an algorithm'' instruction that provides no strategic hints (see \autoref{fig:minimal_prompt}).
\item \textbf{Prompt with high-level direction (``Direction''):} a prompt that gives a high-level strategy: formulate the problem as an optimization and then use approximations to make it tractable (see \autoref{fig:guidance_cloudcast_prompt}).
\item \textbf{Prompt with detailed optimization formulation:} the ``Direction'' prompt augmented with the full mathematical formulation from \autoref{appendix:cloudcast_full_optimization}. 
\end{itemize}

The results will tell us whether performance depends on the creative step of formulating the optimization or the downstream step of engineering tractable approximations.

\begin{figure}[t]
    \centering
    \includegraphics[width=\linewidth]{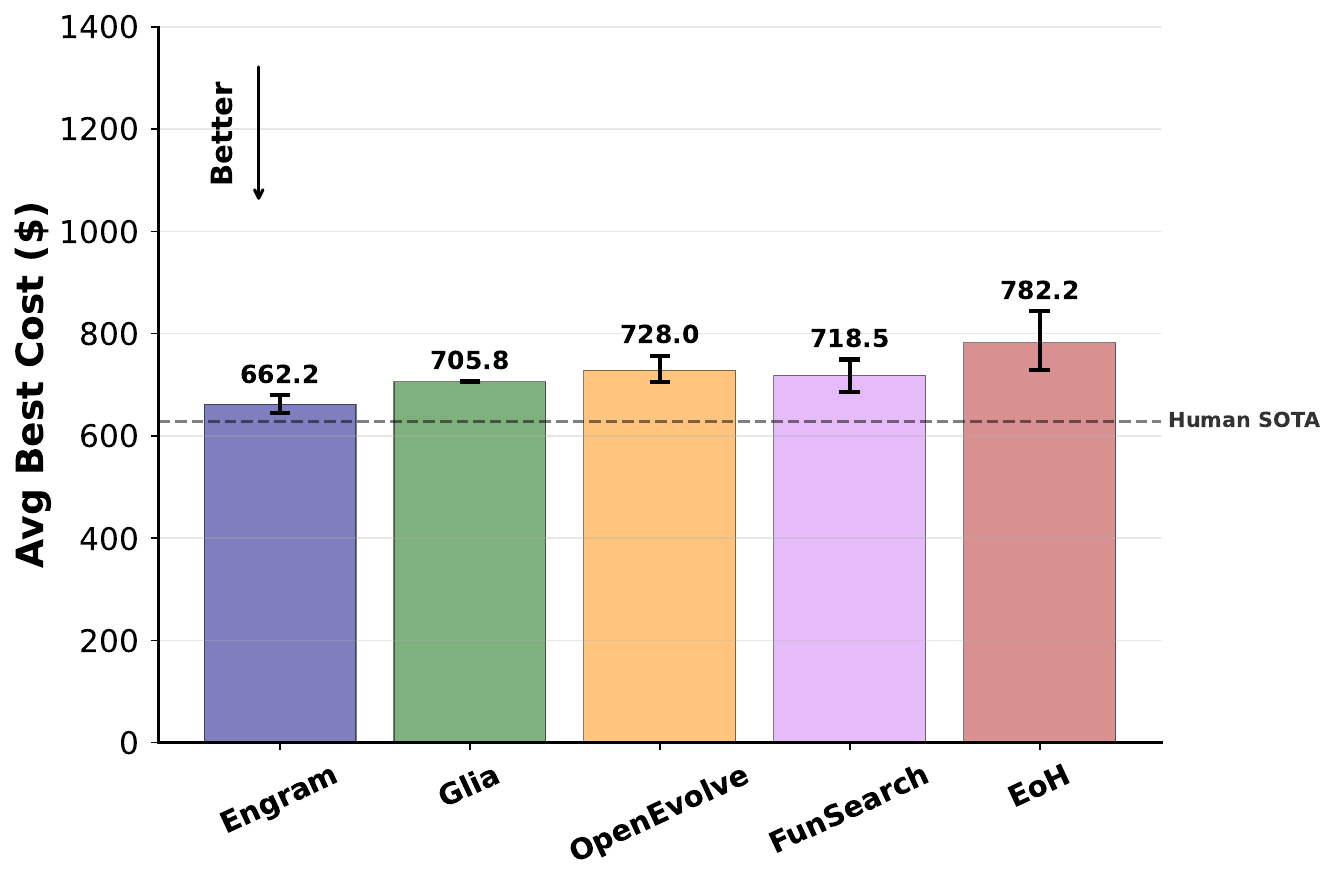}
    \caption{ comparison for multi-cloud multicast with ``Direction'' prompt and \textsc{o3} (lower is better). \name achieves the strongest average best cost, outperforming both evolutionary approaches (EoH, FunSearch, OpenEvolve) and Glia. The whiskers show 90\% confidence intervals.}
    \label{fig:cloudcast_baselines}
\end{figure}

\NewPara{Take-away results.}
Across prompts and models, the strongest performance comes from giving \emph{high-level direction} (\autoref{fig:cloudcast_baselines}): the ``Direction'' prompt consistently yields the lowest costs, while adding the full optimization from \autoref{appendix:cloudcast_full_optimization} produces no measurable improvement. This suggests that \emph{strategic guidance} (``treat this as optimization, then approximate'') matters more than providing the full mathematical formulation, and \name persists within an optimization family until it becomes tractable and succeeds. When removing the direction, evolutionary approaches and Glia remain trapped in Steiner-tree--style~\cite{steiner} heuristics, which typically employ greedy or distance-network approximations. In contrast, \name discovers solver-backed designs that use Dynamic Programming (DP) and Mixed Integer Linear Programming (MILP), which are strong alternatives to the human SOTA. Stronger reasoning models (e.g., \textsc{gpt-5.2}) find these solutions more reliably.

\NewPara{Performance comparison.} Each agent run produces a best-performing candidate.  We call its cost the ``best cost'' for that agent run. We run each agent using the ``Direction'' prompt and the \textsf{o3} LLM 10 times. \autoref{fig:cloudcast_baselines} shows the average and 90\% confidence interval of these 10 best costs. \name outperforms all the other approaches (with average best cost of \$662) and provides solutions closest to human SOTA. It also achieves the strongest LLM-generated result (\autoref{fig:o3-my-handoff-code-cloudcast}) across all the methods and runs, with the best cost reaching \$625 in one agent run, slightly improving human SOTA~\cite{cloudcast} (\$626). \name's best solution uses a tractable Mixed Integer Linear Program (MILP) in the same style as the human SOTA. 


The best EoH/OpenEvolve/FunSearch solutions implement a graph algorithm that builds an approximate Steiner tree, with no explicit optimization attempt, ranging in cost between \$640 and \$696. The best Glia solution is also a graph algorithm without explicit optimization with a cost of \$687, but is better than the other solutions because it comes up with a provider-aware/shared-tree construction (see \autoref{fig:o3-my-glia-code-cloudcast}).

\begin{figure}[t]
    \centering
    \includegraphics[width=\linewidth]{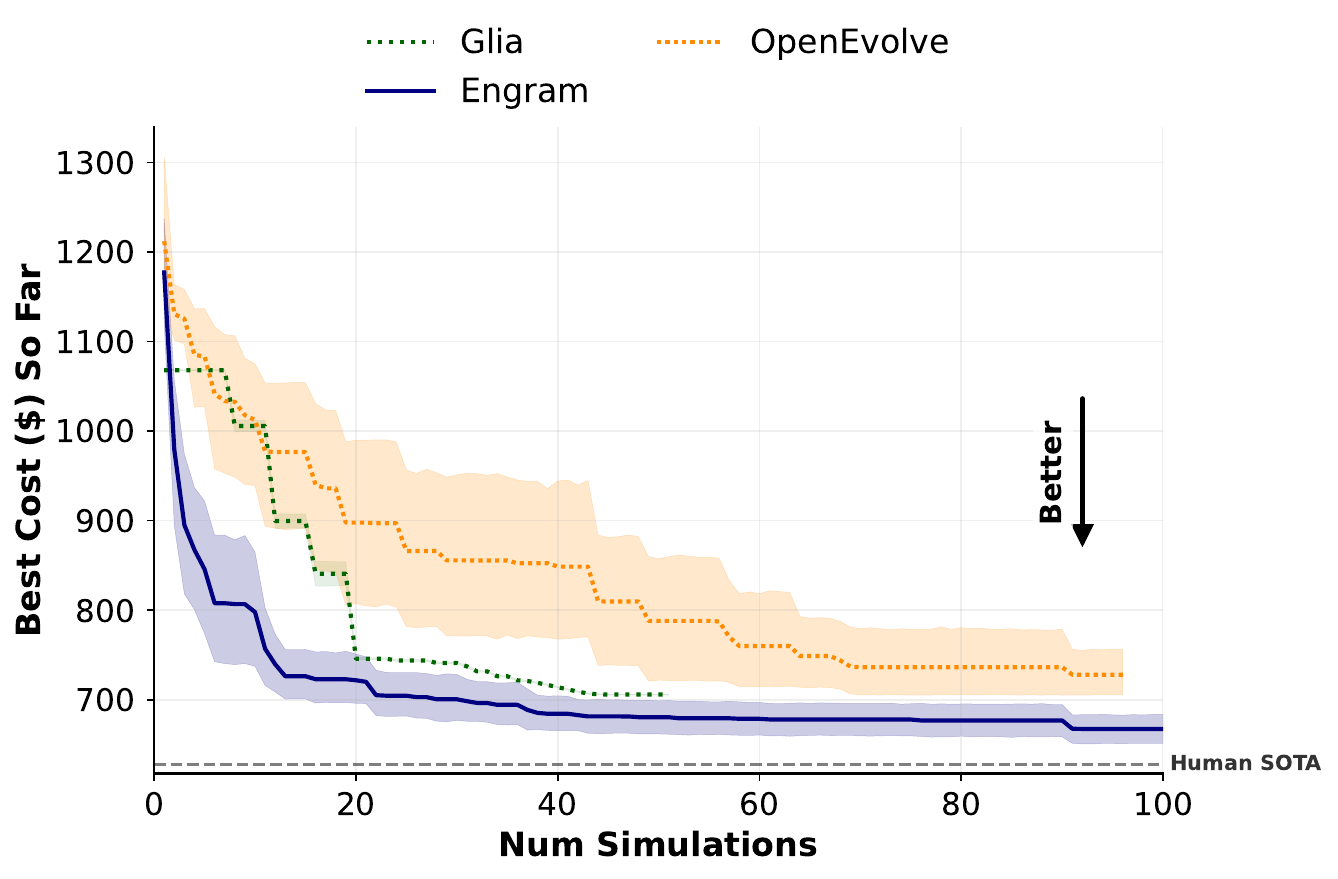}
    \vspace{-10pt}
\caption{Best cost-so-far versus number of simulations on multi-cloud multicast. OpenEvolve plateaus on simple heuristics; Glia makes progress but eventually exhausts its context/budget; \name continues improving and finds solver-based solutions. The shades are 90\% confidence intervals of best costs so far across runs.}

    \label{fig:multicast_motivation}
\end{figure}

\NewPara{Progress results and discussion.}
\autoref{fig:multicast_motivation} (progress curves) shows the best-cost-so-far versus simulation budget for the ``Direction'' prompt and \textsf{o3} model. OpenEvolve initially improves but then plateaus well above Human SOTA. Inspecting the generated heuristics, we find a clear structural pattern: most OpenEvolve~\cite{openevolve} solutions remain variants of Steiner-tree-style backbones with minor local tweaks. This is reminiscent of the neighborhood bias issue with evolutionary approaches discussed in \autoref{s:intro}. OpenEvolve does not progress toward optimization-derived designs. A comment block in one top solution even emphasizes this point explicitly: \emph{``The whole routine is heuristically efficient---no MILP solver invocation---yet it typically cuts total egress cost by $>30\%$.''} (see \autoref{fig:openevolve-code-cloudcast}).

Glia exhibits a different trajectory in \autoref{fig:multicast_motivation}. It shows a sharp improvement initially, but progress eventually stalls despite an available simulation budget due to context limits (\autoref{fig:multicast_motivation}). Progress stalls because of the coherence ceiling discussed in \autoref{s:intro}.

\begin{figure}[t]
\centering

\begin{subfigure}{\linewidth}
    \centering
    \includegraphics[width=0.95\linewidth]{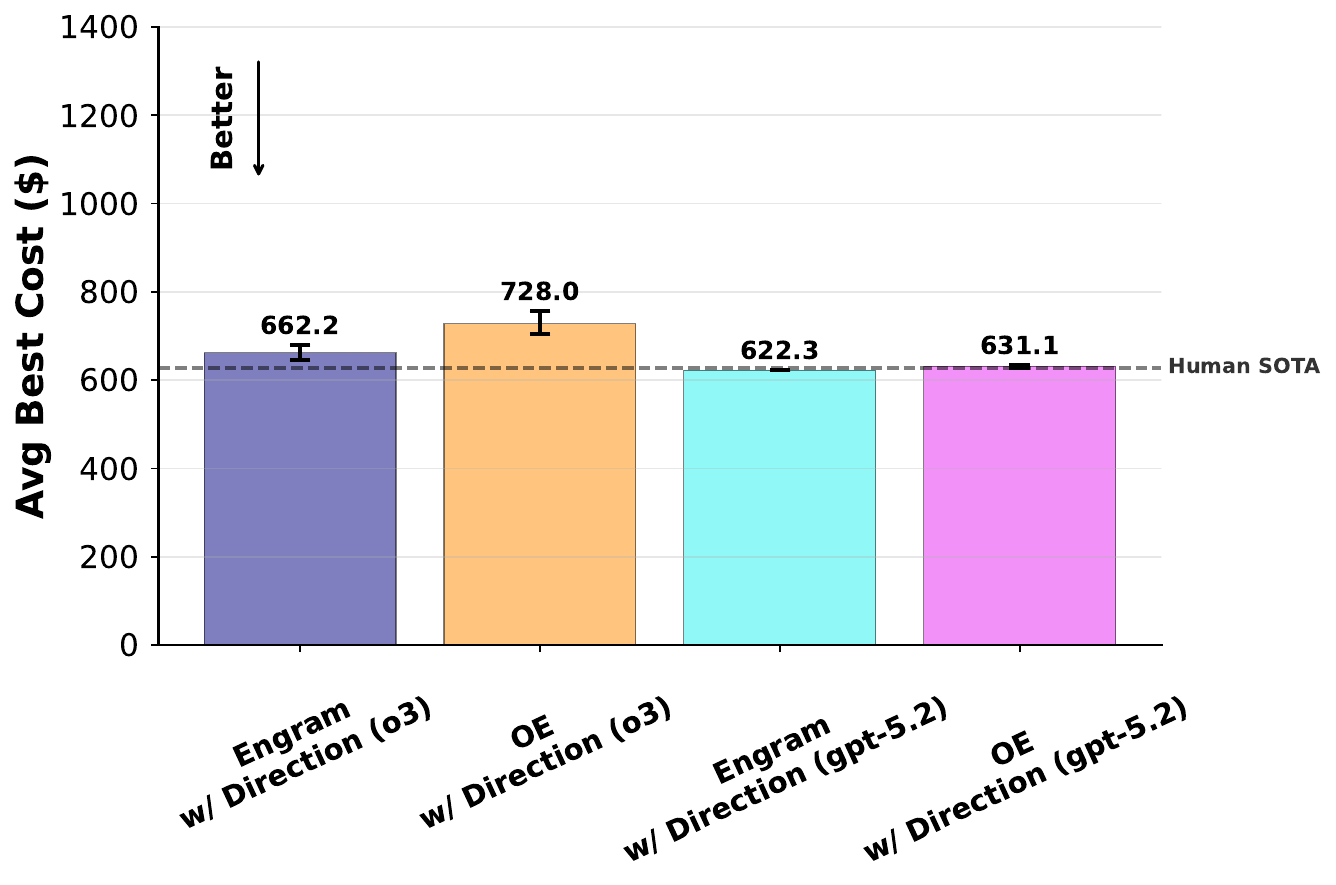}
    \vspace{-10pt}
    \caption{Method comparison with direction prompt.}
    \label{fig:multicast_direction}
\end{subfigure}

\begin{subfigure}{\linewidth}
    \centering
    \includegraphics[width=0.95\linewidth]{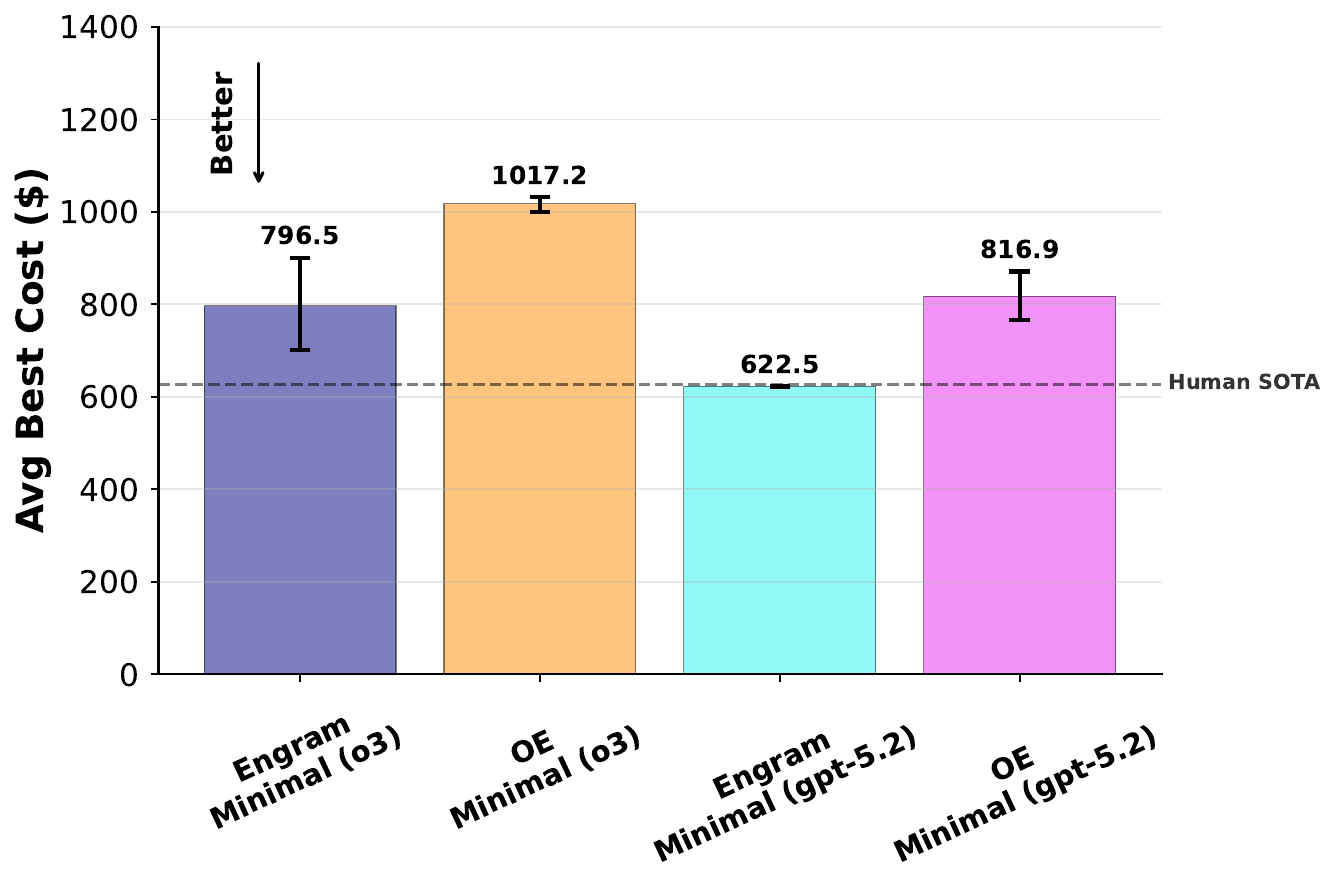}
    \vspace{-10pt}
    \caption{Method comparison with simple minimal prompt.}
    \label{fig:multicast_simple}
\end{subfigure}

\caption{Average and 90\% confidence interval of best-cost 10 runs for multi-cloud multicast (OE: OpenEvolve, lower is better). \name outperforms OpenEvolve in all settings.}
\label{fig:multicast_comparison}
\end{figure}

\NewPara{Prompt and model sensitivity.}
\autoref{fig:multicast_direction} shows the average best costs with ``Direction'' prompt for \textsf{gpt-5.2} as well. For both models, \name rapidly shifts into explicit optimization. OpenEvolve only formulates an optimization in \textsf{gpt-5.2}; with \textsf{o3}, it stays in Steiner-tree-based heuristics. We did not observe any measurable behavior change when we additionally provided the full optimization (\autoref{appendix:cloudcast_full_optimization}); performance and qualitative solution structure remained essentially unchanged.

\autoref{fig:multicast_simple} shows the results using the minimal ``write an algorithm'' prompt for both \textsf{o3} and \textsf{gpt-5.2}. OpenEvolve cannot produce good solutions with this prompt, even with \textsf{gpt-5.2}. However, \name with \textsf{o3} on average reaches better solutions than OpenEvolve, even when OpenEvolve uses \textsf{gpt-5.2}. Surprisingly, in one run, \name generated a heuristic that uses MILP even with the minimal prompt and beats the human SOTA (\$623 vs.\ \$626 for the cost of multicast) (see \autoref{fig:o3-simple-handoff-code-cloudcast}). Moreover, with the stronger model (\textsf{gpt-5.2}), \name introduces a fundamentally different and novel algorithmic approach using dynamic programming (DP) that consistently beats the human SOTA (\$622 vs.\ \$626) (see~\autoref{fig:5.2-simple-handoff-code-cloudcast}). 

\begin{figure}[t]
\centering
\begin{tikzpicture}
\begin{axis}[
  width=\linewidth,
  height=5cm,
  xmin=0.5, xmax=4.5,
  ymin=550, ymax=1225,
  enlarge x limits=0.15,
  enlarge y limits=0.08,
  clip=false,   
  ymajorgrids=true,
  grid style={black!10},
  xlabel={Time},
  ylabel={Cost (\$)},
  xtick={1,2,3,4},
  xticklabels={},
  tick label style={font=\small},
  label style={font=\small},
  axis line style={black!40},
]

\addplot[
  thick,
  mark=*,
  mark size=2.5pt,
] coordinates {
  (1,772)
  (2,1104)
  (3,1104)
  (4,644)
};

\node[font=\small,fill=white,inner sep=2pt]
  at (axis cs:1,772) [above left=2pt]
  {{\color{blue}\textbf{(1) \$772}}};

\node[font=\small,fill=white,inner sep=2pt]
  at (axis cs:2,1104) [above=4pt]
  {\textbf{\textcolor{red}{(2) \$1104}}};

\node[font=\small,fill=white,inner sep=2pt]
  at (axis cs:3,1104) [above=4pt]
  {\textbf{\textcolor{red}{(3) \$1104}} };

\node[font=\small,fill=white,inner sep=2pt]
  at (axis cs:4,644)  [below right=2pt] {\textbf{\textcolor{green!80!black}{(4) \$644}} };

\draw[-{Stealth[length=2mm]},thick,black]
  (axis cs:3,1104) -- 
  node[right,fill=none,inner sep=2pt,font=\small]{recovery}
  (axis cs:4,644);

\draw[-{Stealth[length=2mm]},thick]
  (axis cs:1,772) -- (axis cs:2,1104);

\draw[-{Stealth[length=2mm]},thick]
  (axis cs:2,1104) -- (axis cs:3,1104);

\draw[-{Stealth[length=2mm]},thick]
  (axis cs:3,1104) -- (axis cs:4,644);

\end{axis}
\end{tikzpicture}
\vspace{-10 pt}
\caption{\name tolerates non-monotonic intermediate outcomes while staying within an optimization-based approach. Early attempts temporarily degrade cost ($772 \rightarrow 1104$) before \name recovers and achieves a large improvement ($1104 \rightarrow 644$).}

\label{fig:nonmonotone-cost}
\end{figure}
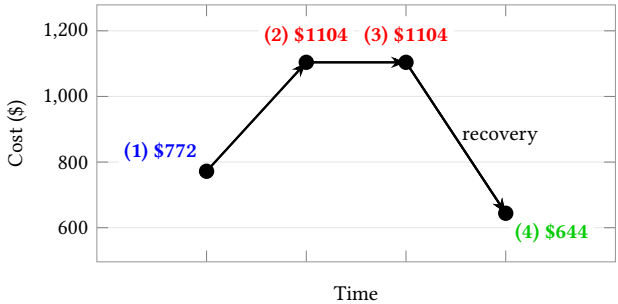

\NewPara{Persistence despite costs sometimes worsening in \name.}
A concrete example shows up in the \textsf{o3}+``Direction'' run (\autoref{fig:nonmonotone-cost}), where \name stays on an optimization approach even though performance regresses before it gets better. {\color{blue}\textbf{(1)}} Starting from a strong baseline (cost \$772), the next agent tries to implement the body of an optimization \textbf{\textcolor{red}{(2)}} and the cost \emph{``EXPLODED to \$1104''}; {\color{red}\textbf{(3)}} a follow-up attempt fails and finds \emph{``NO opportunities''} to improve the cost. Instead of abandoning the optimization direction and reverting to primitive heuristics, the next agent makes the right call and continues {\color{green!80!black}\textbf{(4)}}. It implements a reduced-edge MILP with explicit tractability knobs. This recovers the heuristic and produces the key jump: \emph{``Reduced-edge MILP v2 ... Achieved total cost \$$644$ ($-17$\% vs previous \$$772$ baseline)''} with a cost decrease from $772 \rightarrow 644$. \name tolerates temporary score degradation, uses the failure to diagnose what is missing, and then continues refining within the same algorithmic family until the optimization becomes tractable and wins.

\NewPara{Summary.}
Multi-cloud multicast exposes a regime where the design space is combinatorially large and effective solutions require \emph{optimization structure} rather than local heuristic tweaks. In this setting, giving the model high-level direction (``formulate an optimization, then approximate'') is sufficient to unlock strong performance; providing the full mathematical formulation does not materially change outcomes. \name is uniquely effective at escaping Steiner-tree--style neighborhood bias, persisting through non-monotonic intermediate results, and ultimately producing solver-backed designs (MILP/DP) that match and in some cases surpass Human SOTA.

\subsection{Case Study: LLM Request Routing}
\label{sec:eval-vidur}
\NewPara{Problem overview.} We study the problem of routing requests among replicated LLM instances in distributed serving systems. We evaluate how \name identifies novel routing strategies that lower overall mean request completion time (RT).
We evaluate using \texttt{vidur}, a simulator for distributed LLM serving systems~\cite{agrawal2024vidur}. The main metric is mean request completion time (RT), i.e., end-to-end response latency. RT reflects end-user's perceived quality of responsiveness.

\NewPara{Benchmark.} We use the same setting as Glia~\cite{glia}. We simulate a ShareGPT-based~\cite{sharegpt} LLM inference workload on four NVIDIA A10 GPUs serving Llama-3-8B-Instruct. To mimic reasoning-heavy workloads with heavy-tailed sequence lengths, we independently increase 5\% of prompt lengths and 5\% of decode lengths by 10$\times$. The system operates at 7.5 queries per second (QPS), with bursty arrivals drawn from a log-normal distribution ($\sigma = 2$). Each replica uses chunked prefill~\cite{sarathi} with a chunk size of 8192 for batch scheduling. All experiments are repeated across ten random seeds.

\NewPara{Baseline routing comparisons.} We evaluate \name against three standard routing heuristics: Round-Robin, Least-Loaded Queue (LLQ), and Least Outstanding Requests (LOR). Round-Robin cycles requests across replicas; LLQ selects the replica with the fewest active requests; and LOR chooses the replica with the fewest queued requests awaiting GPU allocation. We also compare against a workload-specialized heuristic designed over two weeks by a senior systems researcher with over two decades of experience~\cite{glia}.

\begin{figure}
    \centering
    \includegraphics[width=\linewidth]{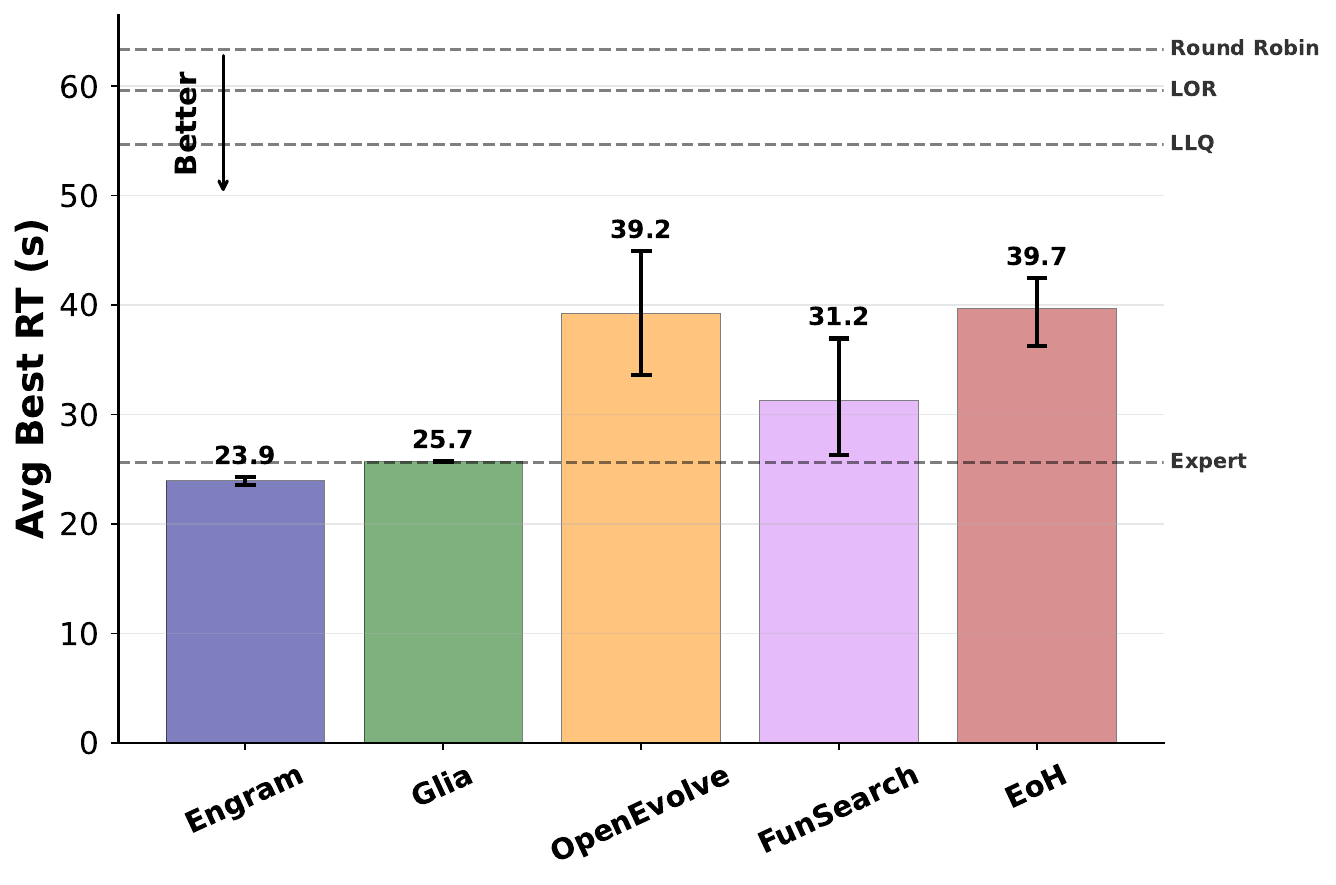}
    \caption{Best-response time comparison on the LLM request routing problem (lower is better). \name achieves the strongest average best response time, outperforming evolutionary approaches (EoH, FunSearch, OpenEvolve), Glia, and Expert. The whiskers show 90\% confidence intervals.}
    \label{fig:vidur_baselines}
\end{figure}

\NewPara{Quantitative and qualitative results.} 
\autoref{fig:vidur_baselines} shows that \name achieves the best aggregate performance among the LLM-based methods, with the lowest average best RT (lower is better). In particular, \name outperforms evolutionary code-search methods (EoH, FunSearch, OpenEvolve) as well as the Glia variants, and outperforms the expert-designed routing heuristic. \name is not only capable of finding strong novel routing policies, but does so consistently across runs (exhibiting low performance variations).

Qualitatively, the strongest policy found by \name is not a simple queue-length heuristic (like LLQ/LOR) (see \autoref{fig:o3-vidur-handoff-code}), but a structured routing strategy that combines \textbf{Shortest-Prefill-First (SPF) request ordering}, \textbf{decode-sensitive per-replica pending-queue caps}, and a \textbf{memory-utilization admission guard} that avoids placements likely to trigger evictions. It then selects among admissible replicas using \textbf{projected memory utilization}, which directly targets the prefill/decode interference pattern induced by chunked prefill and heavy-tailed requests. In other words, \name discovers a policy that explicitly reasons about the interaction between memory pressure and decode-phase contention, rather than optimizing only coarse load counts. Comparing best solutions of Glia and \name, Glia uses SPF and a KV-cache headroom–based memory admission rule~\cite{glia}, but \name goes further by adding decode-sensitive pending-queue caps and decode-aware tie-breaking, making its routing policy more explicitly responsive to decode-phase contention rather than primarily enforcing memory safety through reserved free blocks.

By comparison, the best evolutionary solution we observed remains closer to a generic shortest job first heuristic: it also uses prefill-aware ordering and predicted work estimates, but primarily scores replicas using a token-based outstanding-work proxy with a fixed memory guard.

\subsection{Case Study: Optimizing KV Cache Reuse in Databases with Natural Language Queries}
\label{eval-llmsql}
This problem  addresses the cost of batch LLM inference over relational tables \cite{liu2025optimizing}. When an LLM processes rows sequentially, consecutive rows that share a long serialized prefix can reuse the key–value (KV) cache, thereby reducing inference cost. Liu et al.~\cite{liu2025optimizing} propose a SOTA method for reordering a dataframe to maximize prefix-based KV-cache reuse. Their key insight is that allowing per-row dynamic field ordering, rather than enforcing a fixed column order across all rows, can substantially improve cache hit rates. This observation motivated the design of their algorithm, Greedy Group Recursion (GGR). Note that a brute-force search over all possible reorderings is computationally infeasible. For a table with $m$ rows and $n$ columns, the total number of possible reorderings is prohibitively large, $n!\times (m!)^n$~\cite{liu2025optimizing}.

\begin{figure}[t]
    \centering
\includegraphics[width=0.9\linewidth]{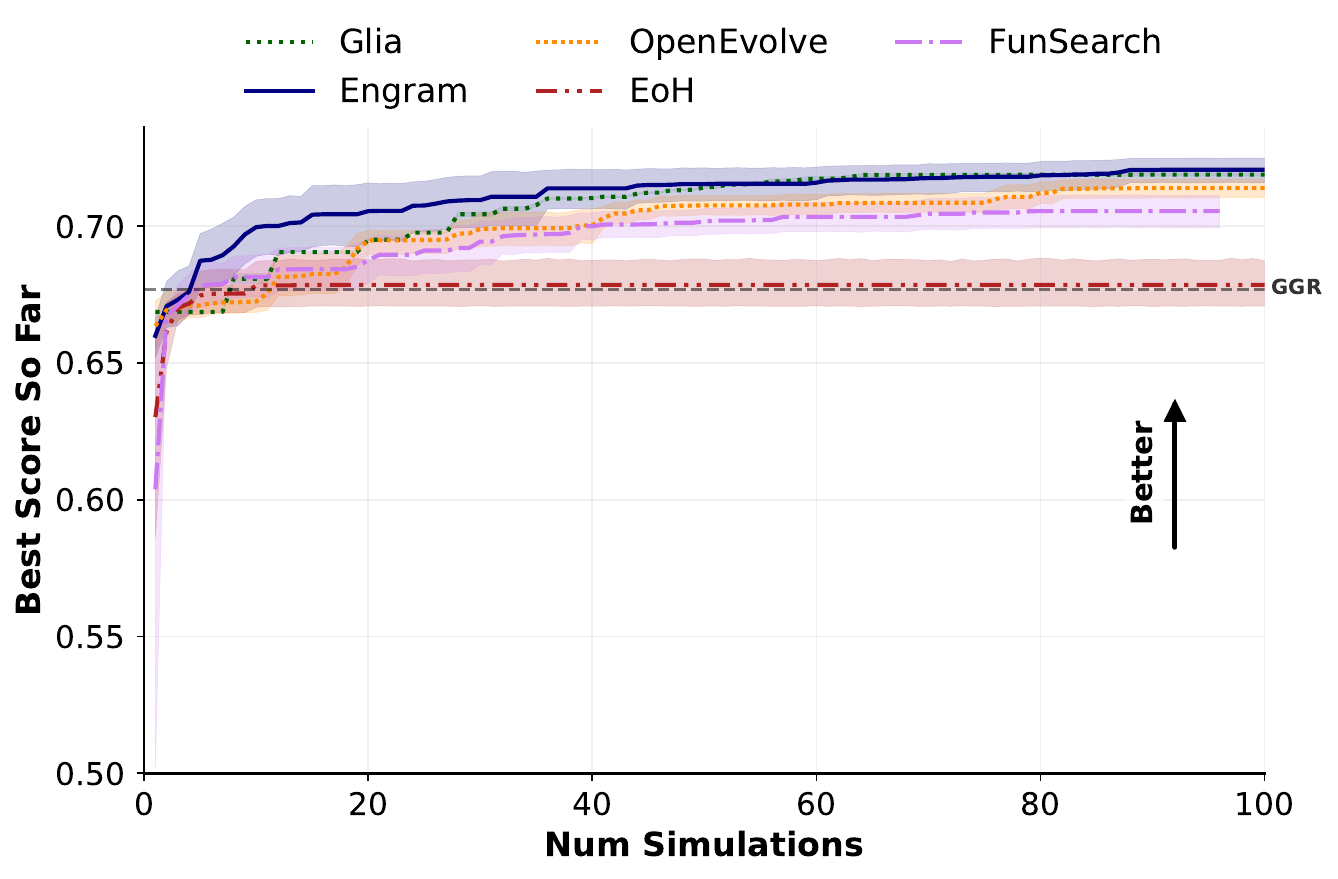}
    \vspace{-10pt}
    \caption{Average best score so far vs.\ number of simulations for all methods seeded with a simple baseline program for the LLM-SQL task. \name, Glia, and OpenEvolve converge to comparable final scores with \name being faster in convergence, while FunSearch and EoH lag behind. The GGR \cite{liu2025optimizing} line shows the score of the state-of-the-art algorithm.}
    \label{fig:llmsql_normal}
\end{figure}

We adopt the evaluation environment of ADRS \cite{cheng2025let}, running each algorithm on five datasets (\textsc{movies}, \textsc{beer}, \textsc{bird}, \textsc{pdmx}, and \textsc{products}) and reporting the compined score of  $0.95 \times \bar{h} + 0.05 \times r$, where $\bar{h}$ is the mean per-dataset prefix hit rate and $r$ is a normalized runtime bonus capped at 12 seconds. In addition, since we observed that pre-merging columns with functional dependencies benefits all methods, we apply this merge before reordering for all algorithm at evaluation time rather than leaving it to each algorithm. 

We seed all the methods with a simple baseline program that sorts columns by cardinality and orders rows lexicographically. We use a task prompt (\autoref{fig:sql-simple-prompt}), which describes the full problem, indicating the benefits of having different column-ordering per-row. Under this setup, all methods perform competitively. \name attains the highest mean best score (0.721), followed closely by Glia (0.719) and OpenEvolve (0.714) (see ~\autoref{fig:llmsql_normal}). \name is the most efficient method, requiring fewer simulations to achieve any given score.

Interestingly, the best programs produced by all methods did not use GGR's recursive structure, yet still achieved high scores. These high-scoring non-recursive methods all leveraged per-row column reordering to improve cache reuse.

For this case study, we additionally evaluate \name and OpenEvolve when initialized with the GGR algorithm instead of the simple baseline. Experimental details are provided in \autoref{app: llm-sql}.
\begin{table*}[t]
\centering
\small
\setlength{\tabcolsep}{8pt}
\renewcommand{\arraystretch}{1.25}
\begin{tabular}{lcccccc}
\toprule
\textbf{Strategy} 
& \textbf{CBL $\downarrow$} 
& \textbf{CBL-Multi $\downarrow$} 
& \textbf{EPLB $\uparrow$} 
& \textbf{Prism $\uparrow$} 
& \textbf{Telemetry $\uparrow$} 
& \textbf{TXN $\uparrow$} \\
\midrule
Human SOTA 
& 101.7 
& 92.3 
& 0.251 
& 21.89 
& 0.822 
& 2724.8 \\

\midrule
\name
& 103.6 $\pm$ 1.1 
& \textbf{79.9 $\pm$ 0.8} 
& \textbf{0.273 $\pm$ 0.00} 
& \textbf{27.94 $\pm$ 1.70} 
& \textbf{0.954 $\pm$ 0.00} 
& \textbf{3918.6 $\pm$ 56.6} \\

OE~\cite{openevolve}
& 103.4 $\pm$ 0.9 
& 79.9 $\pm$ 0.4 
& 0.214 $\pm$ 0.06 
& 26.21 $\pm$ 0.03 
& 0.953 $\pm$ 0.00 
& 3713.7 $\pm$ 77.9 \\
\bottomrule
\end{tabular}
\caption{Average best scores across 10 runs with 90\% confidence intervals (higher is better $\uparrow$; lower is better $\downarrow$). \name exceeds Human SOTA on five of six tasks and improves over OpenEvolve on four.}
\label{tab:adrs_large}
\end{table*}

\section{Additional Evaluation}
\label{sec:eval_large}

We evaluate \name and OpenEvolve~\cite{openevolve} on a larger set of ADRS problems~\cite{UCB_ADRS_GitHub, cheng2025let} and compare to the reported Human SOTA. For cost, we use OpenAI \textsf{o3}~\cite{openai_o3_o4_mini_system_card}. Each method runs 10 times; we report average best scores with 90\% confidence intervals (Table~\ref{tab:adrs_large}). \name exceeds Human SOTA on five of six tasks and improves over OpenEvolve on four. In \autoref{sec:auc}, we show that \name is faster than OpenEvolve in reaching its peak performance.
\begin{figure}
    \centering
    \includegraphics[width=\linewidth]{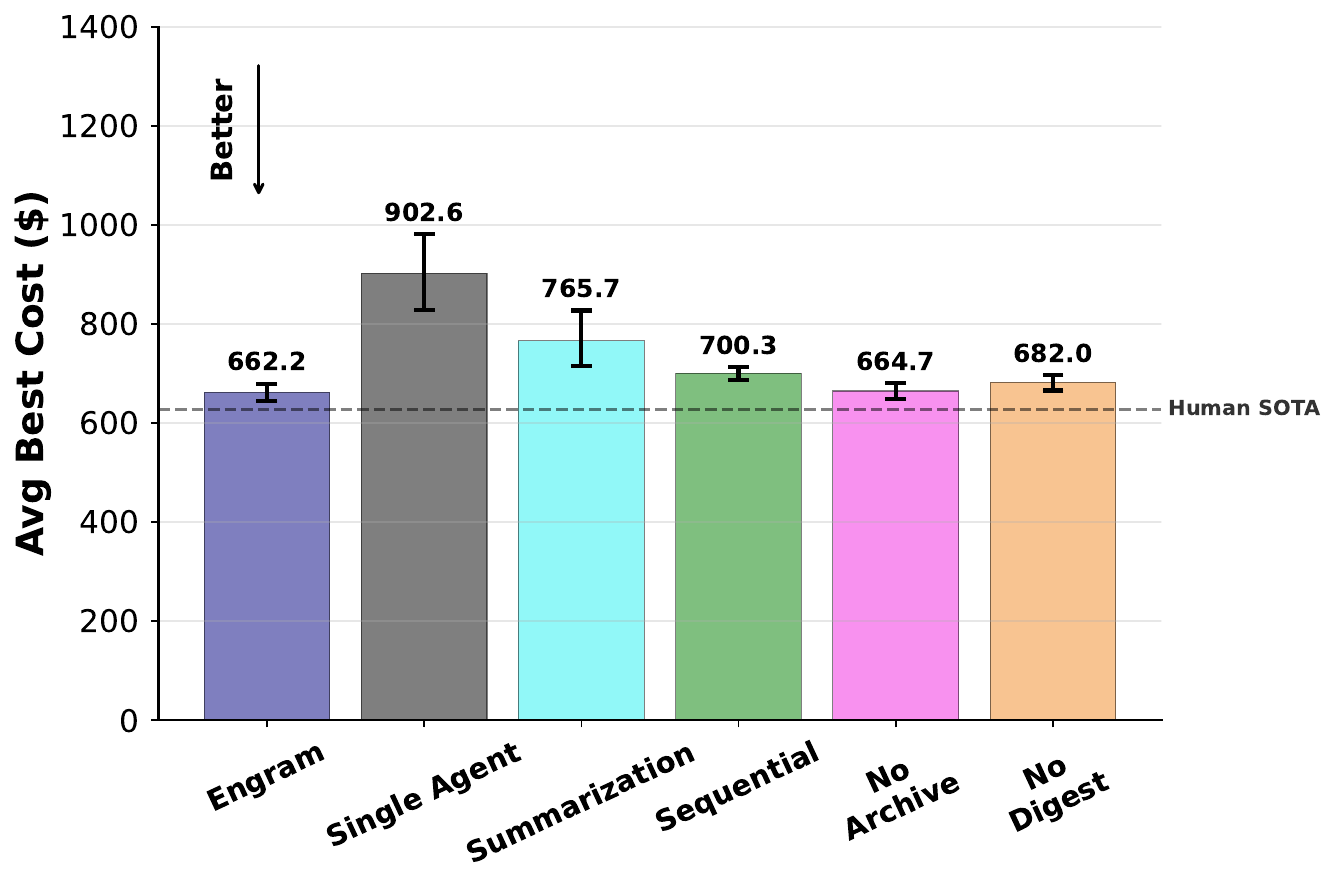}
    \caption{Ablation on multi-cloud multicast (average best cost; lower is better). Single-agent variants (Single Agent and Summarization) are constrained by context growth and perform worst, while multi-agent variants improve upon them. \name performs better than mere Sequential code transfer, highlighting the benefit of persistent knowledge. Removing either the \archive or Digest slightly worsen the performance.}
    \label{fig:cloudcast_ablation}
    \vspace{-10 pt}
\end{figure}

\NewPara{Ablation experiments.}
\label{sec:ablation} We ablate the main components of \name on the multi-cloud data transfer task (\autoref{fig:cloudcast_ablation}) to isolate the effects (i) coherence ceiling, and (ii) persistent cross-agent knowledge transfer.

The \textbf{Single Agent} is an agent with full tool and execution access but no other structure. Single Agent performs worst (average cost \$902); it exhausts its effective reasoning budget and runs fewer meaningful experiments, and converges to weaker heuristics. \textbf{Summarization} improves upon the single agent (average cost \$765) and tries to enable a longer context by and compressing older context and allowing additional iterations.
However, it long-horizon coherence remains limited and confined to a single summarized context and still underperforms all multi-agent variants.

In \textbf{Sequential} variant, the best code from one agent is passed to the next without any structured context sharing (no \ledger and no \archive). This isolates the benefit of simply resetting context across agents. Sequential improves performance (average cost \$700) relative to single-agent baselines, showing that fresh contexts help. However, it still falls short of full variations with persistent cross-agent knowledge transfer.

The \textbf{No Archive} and \textbf{No Digest} variants each remove one component of \name, leading to modest but consistent performance degradation compared to the full system. Removing the Digest has a larger negative impact, suggesting that Digest plays a more critical role in guiding future agents than raw information in \archive.

\section{Other Related Work}
\label{sec:related}


LLMs are increasingly used for algorithm discovery via \emph{reasoning-and-search} loops rather than one-shot code generation~\cite{liu2024llm4ad,xie2025far}. Prior systems span (i) multi-agent iterative refinement~\cite{gottweis2025towards}, (ii) evolutionary and tree-search methods that optimize populations of programs using fitness-based selection (e.g., EoH, ShinkaEvolve, AlphaEvolve, MCTS, LAS, X-evolve)~\cite{EoH,lange2025shinkaevolve,AlphaEvolve,zheng2025monte,liu2025fitness,zhai2025x}, including recent multi-objective variants~\cite{yao2025multi}, and (iii) approaches that incorporate learning signals such as supervised/fine-tuning on curated traces or datasets~\cite{huang2025calm,liu2025fine}. Recent agentic tree-search pipelines and extensions that ground proposals via web research have also been explored~\cite{aygun2025ai,liu2025scientific}. Tool-augmented long-horizon discovery has been studied in symbolic regression as well (e.g., SR-Scientist)~\cite{xia2025sr}.

\paragraph{LLMs for systems research.}
A growing body work is emerging to automate systems research with LLMs. At a high level, these efforts span broad systems domains (e.g., ADRS)~\cite{cheng2025barbariansgateaiupending} as well as targeted optimization of performance-critical artifacts such as C++ code~\cite{shypula2025automated} and numerical kernels (e.g., AlgoTune)~\cite{press2025algotune}. Several works formalize this interaction through specialized interfaces and feedback interpreters: Wei et al.~\cite{wei2024improving} propose an Agent-System Interface (ASI) built around a compact DSL plus an AutoGuide feedback layer to optimize mapper code for parallel programs, while related ASI-style closed loops have been extended to other design spaces such as neural architectures (ASI-ARCH)~\cite{liu2025alphago}. Glia~\cite{glia} replaces direct code mutation of prior work with reasoning-based exploration following the scientific method.

LLM-driven search has also been applied to discovering or tuning algorithms in diverse problem settings, including SAT solving~\cite{sun2025automatically}, adaptive bitrate streaming~\cite{10.1145/3696348.3696868}, and networking control policies such as congestion control and caching~\cite{he2025congestioncontroloptimizationlarge,dwivedula2025man}. In performance engineering for accelerators, Astra~\cite{wei2025astra} and GPU Kernel Scientist~\cite{andrews2025gpu} study LLM-guided optimization of CUDA kernels. Beyond systems, similar search-and-improve templates appear in theoretical and combinatorial discovery, e.g., using AlphaEvolve to find new constructions in complexity theory~\cite{nagda2025reinforced}.

Multiple works emphasize structured exploration to improve both quality and novelty. Robusta~\cite{karimi2025robusta} combines combinatorial reasoning from prior work~\cite{10.1145/3696348.3696884,metaease} with LLM-guided evolutionary search to obtain networking heuristics with stronger worst-case guarantees, while MetaMuse~\cite{ma2025algorithm} steers generation using external stimuli, waypoint reasoning, and feedback-derived performance embeddings to promote diverse, high-performing algorithms (e.g., cache replacement and bin packing).

\section{Conclusion}
\label{sec:conclusion}

This paper introduced \name, an agentic LLM researcher architecture designed to improve coherence in long-horizon system design tasks. We identified two core limitations of prior LLM-based approaches: evolutionary neighborhood bias and the coherence ceiling. \name addresses these limitations by decoupling long-horizon exploration from the constraints of a single context window. It organizes exploration into a sequence of agents that iteratively design, test, and analyze mechanisms. At the conclusion of each run, an agent stores code snapshots, logs, and results in a persistent {\em \archive} and distills high-level modeling insights into a compact, persistent {\em \Ledger}. Subsequent agents then begin with a fresh context window, reading the \Ledger to build on prior discoveries. 

Across three diverse case studies, \name consistently outperformed evolutionary and iterative agentic baselines. Beyond improved performance, our results show that \name can navigate conceptual boundaries more effectively, tolerate temporary regressions, persist within promising algorithmic families, and produce innovative and principled designs.

\section*{Acknowledgments}
This work was partially funded by the MIT Generative AI Impact Consortium (MGAIC) and Quanta Computer, Inc. under the AIR Project. We would like to thank Diego de Lope and Avi Kapur Srinivasan for their valuable insights.

\newpage
\bibliographystyle{ACM-Reference-Format}

\newpage
\clearpage
\appendix

\section{Appendix}

\subsection{Formulation for multi-cloud multicast}
\label{appendix:cloudcast_full_optimization}

We provide the classical mixed-integer optimization formulation underlying the multi-cloud multicast.
The goal is to design a multicast delivery plan that jointly optimizes monetary cost and feasibility under
a completion-time budget, by deciding: (i) where to place relay capacity (VMs) in regions, (ii) how to route
traffic over heterogeneous inter-cloud links, and (iii) how to respect per-VM ingress/egress limits.

\paragraph{Topology and inputs.}
Let $G=(V,E)$ be a directed graph of cloud regions $V$ and directed transfer links $E$.
A transfer instance specifies a source region $s\in V$, a destination set $D\subseteq V$,
a total transfer size $\mathsf{B}$ (GB), a completion-time budget $\mathsf{T}$ (seconds), and a number
of \emph{stripes} (partitions) $\mathsf{K}\in \mathbb{Z}_{\ge 1}$. Each stripe has volume
$\mathsf{b}=\mathsf{B}/\mathsf{K}$.

Each directed edge $(u,v)\in E$ has:
(i) an egress price $c_{u,v}$ (\$/GB), and
(ii) a profiled per-VM throughput $\beta_{u,v}$ (GB/s) achievable from $u$ to $v$.
Each region $v$ has a per-VM egress limit $e_v$ (GB/s), a per-VM ingress limit $i_v$ (GB/s),
a per-VM instance price $p_v$ (\$/s), and a maximum VM count $\ell_v$.

\paragraph{Decision variables (relay placement and routing).}
Cloudcast uses three sets of variables:
\begin{itemize}
  \item \textbf{Stripe routing / tree selection:} $P_{k,(u,v)}\in\{0,1\}$ indicates whether stripe $k\in\{1,\dots,\mathsf{K}\}$
  is sent over edge $(u,v)\in E$. The selected edges form a multicast distribution structure for each stripe.
  \item \textbf{Relay/VM placement:} $N_v\in\mathbb{Z}_{\ge 0}$ is the number of VMs (overlay routers) provisioned in region $v$.
  This captures \emph{relay placement and capacity}: regions with $N_v>0$ can originate/forward traffic subject to limits.
  \item \textbf{Auxiliary connectivity/acyclicity flow:} $F_{k,(u,v)}\in\mathbb{R}_{\ge 0}$ is an auxiliary flow used only to enforce that,
  for each stripe, the selected edges are connected and cycle-free and allow flow from the source to all destinations.
\end{itemize}

\paragraph{Objective (instance cost + egress cost).}
The objective minimizes the sum of (i) VM instance cost over the time budget and (ii) egress cost for the
data volume shipped along selected edges:
\begin{align}
\min_{P,N,F}\quad
\mathsf{T}\sum_{v\in V} p_v N_v
\;+\;
\frac{\mathsf{B}}{\mathsf{K}}
\sum_{k=1}^{\mathsf{K}}\sum_{(u,v)\in E} c_{u,v}\, P_{k,(u,v)}.
\label{eq:cloudcast_obj}
\end{align}

\paragraph{VM count limits (relay placement constraints).}
\begin{align}
\forall v\in V:\quad N_v \le \ell_v.
\label{eq:cloudcast_vm_limit}
\end{align}

\paragraph{Completion-time feasibility via volume capacities.}
To enforce completion within $\mathsf{T}$ while keeping the program linear, Cloudcast expresses bandwidth constraints
as \emph{volume capacities} over the time horizon and allocates volume in stripe units.

\emph{(i) Edge capacity constraints.}
The total stripe volume placed on edge $(u,v)$ must fit in the volume that $N_u$ VMs at $u$ can transmit in time $\mathsf{T}$:
\begin{align}
\forall (u,v)\in E:\quad
\frac{\mathsf{B}}{\mathsf{K}} \sum_{k=1}^{\mathsf{K}} P_{k,(u,v)}
\;\le\;
\mathsf{T}\, N_u\, \beta_{u,v}.
\label{eq:cloudcast_edge_capacity}
\end{align}

\emph{(ii) Per-region egress constraints.}
A region's aggregate outgoing stripe volume is limited by its per-VM egress cap:
\begin{align}
\forall v\in V:\quad
\frac{\mathsf{B}}{\mathsf{K}} \sum_{k=1}^{\mathsf{K}}\sum_{(v,u)\in E} P_{k,(v,u)}
\;\le\;
\mathsf{T}\, N_v\, e_v.
\label{eq:cloudcast_region_egress}
\end{align}

\emph{(iii) Per-region ingress constraints.}
Similarly, aggregate incoming stripe volume is limited by per-VM ingress:
\begin{align}
\forall v\in V:\quad
\frac{\mathsf{B}}{\mathsf{K}} \sum_{k=1}^{\mathsf{K}}\sum_{(u,v)\in E} P_{k,(u,v)}
\;\le\;
\mathsf{T}\, N_v\, i_v.
\label{eq:cloudcast_region_ingress}
\end{align}

\paragraph{Ensuring a valid multicast structure (connectivity and acyclicity).}
Following Cloudcast, we augment the graph with a special sink node $t$ that is connected only from destinations
(via virtual edges $(d,t)$ for $d\in D$; these incur no cost and are used only for feasibility constraints).
We introduce an auxiliary flow variable $F_{k,(u,v)}\in\mathbb{R}_{\ge 0}$ that enforces that the edges selected by
$P_{k,(u,v)}$ form a connected distribution structure that reaches all destinations for each stripe $k$.

\emph{(i) If an edge is selected, it must carry flow.}
\begin{align}
\forall k,\forall (u,v)\in E':\quad F_{k,(u,v)} \ge P_{k,(u,v)}.
\label{eq:cloudcast_F_lower}
\end{align}

\emph{(ii) Upper bounds on flow (including sink constraints).}
\begin{align}
\forall k,\forall (u,v)\in E':\quad
F_{k,(u,v)} \le
\begin{cases}
1, & u\in D,\ v=t,\\
0, & P_{k,(u,v)} = 0,\\
|D|, & \text{otherwise.}
\end{cases}
\label{eq:cloudcast_F_upper}
\end{align}

\emph{(iii) Flow conservation.} For each stripe $k$, enforce that $|D|$ units are pushed from the source to the sink:
\begin{align}
\forall k,\forall v\in V\cup\{t\}:\quad
\sum_{u} F_{k,(u,v)} - \sum_{w} F_{k,(v,w)}
=
\begin{cases}
|D|, & v=s,\\
-|D|, & v=t,\\
0, & \text{o.w.}
\end{cases}
\label{eq:cloudcast_F_cons}
\end{align}

Intuitively, the sink can receive at most one unit from each destination (Eq.~\eqref{eq:cloudcast_F_upper}),
so pushing $|D|$ units into $t$ forces every destination to be connected to $s$ along edges with positive flow,
which in turn implies those edges are selected by $P$ (Eq.~\eqref{eq:cloudcast_F_lower}).

\paragraph{Why it is intractable at realistic scales.}
The MILP’s binary decision tensor $P\in\{0,1\}^{|E|\times \mathsf{K}}$ induces an exponential search space.
As noted by Cloudcast, the formulation has search space size $O(2^{|V|^2\cdot \mathsf{K}})$; for their 71-region,
10-stripe setting this becomes $O(2^{50410})$, which is infeasible to solve within minutes even with advanced solvers.
This motivates structured relaxations/approximations (e.g., node clustering, hop limits, and stripe-iterative solving)
rather than solving the full MILP at scale.

\section{Additional Experiments on the LLM-SQL Case Study}
\label{app: llm-sql}
In this section, we examine how different initialization choices and prompt formulations affect optimization performance in the LLM-SQL case study.

\begin{figure}[t]
    \centering
\includegraphics[width=0.9\linewidth]{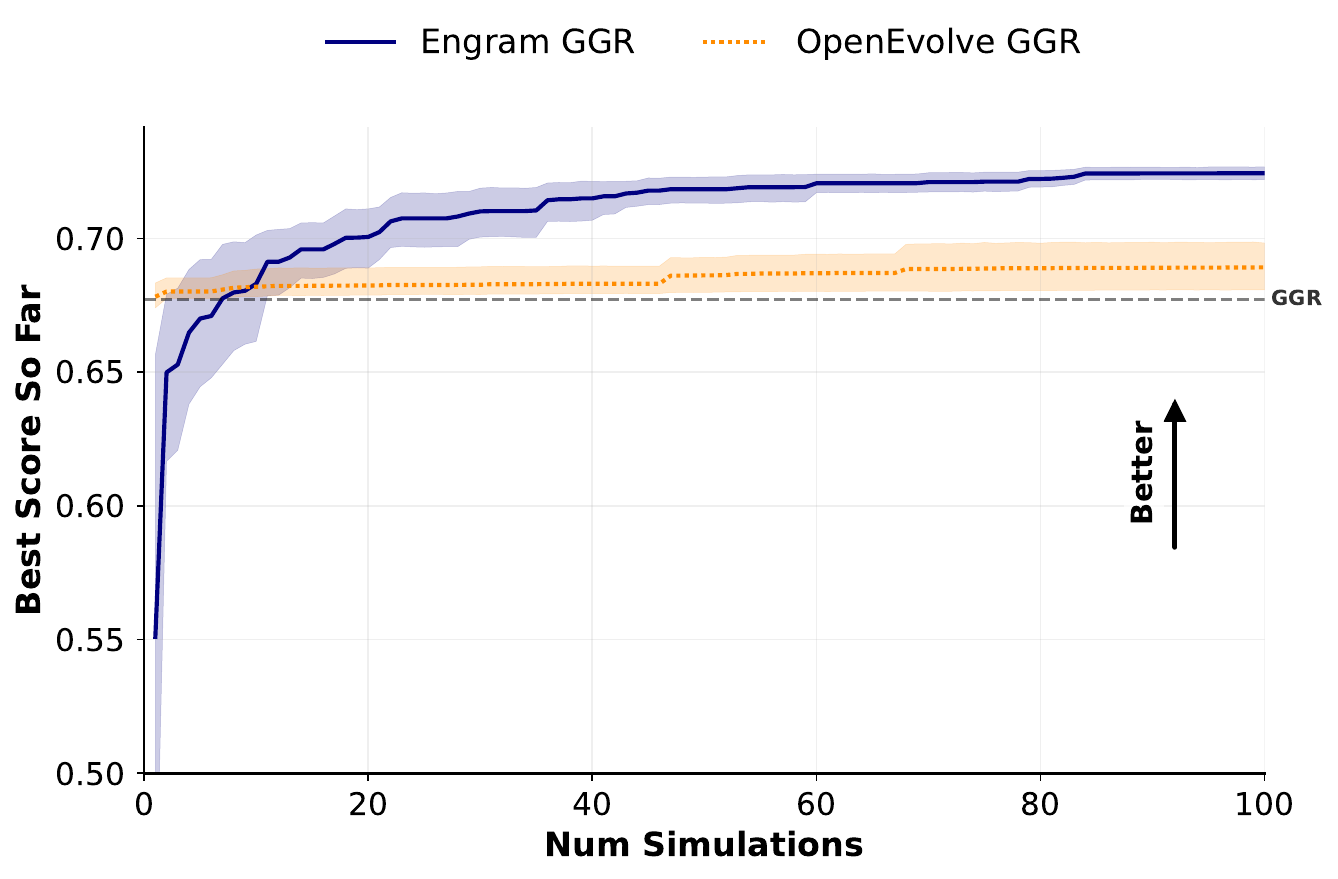}
    \vspace{-10pt}
    \caption{Average score vs.\ number of simulations for all methods seeded with the GGR baseline program for the LLM-SQL task using the same prompt as the Simple case in Figure \ref{fig:llmsql_normal}.}
    \label{fig:llmsql_ggr_ours}
\end{figure}

\NewPara{GGR initialization.} 
To evaluate how well \name and OpenEvolve exploit a high-quality seed, we replace the simple baseline with GGR, the human-designed state-of-the-art algorithm from~\cite{liu2025optimizing}. GGR incorporates nested recursion and thread pooling. Using the same task prompt as in~\ref{eval-llmsql}, the two methods diverge sharply (see~\autoref{fig:llmsql_ggr_ours}). \name achieves an average score of $0.724$, modestly improving over both its performance with the simple baseline and the original GGR implementation. In contrast, OpenEvolve drops to an average of $0.689$, substantially below its score when initialized from the simple baseline.

\begin{figure}[t]
    \centering
\includegraphics[width=0.9\linewidth]{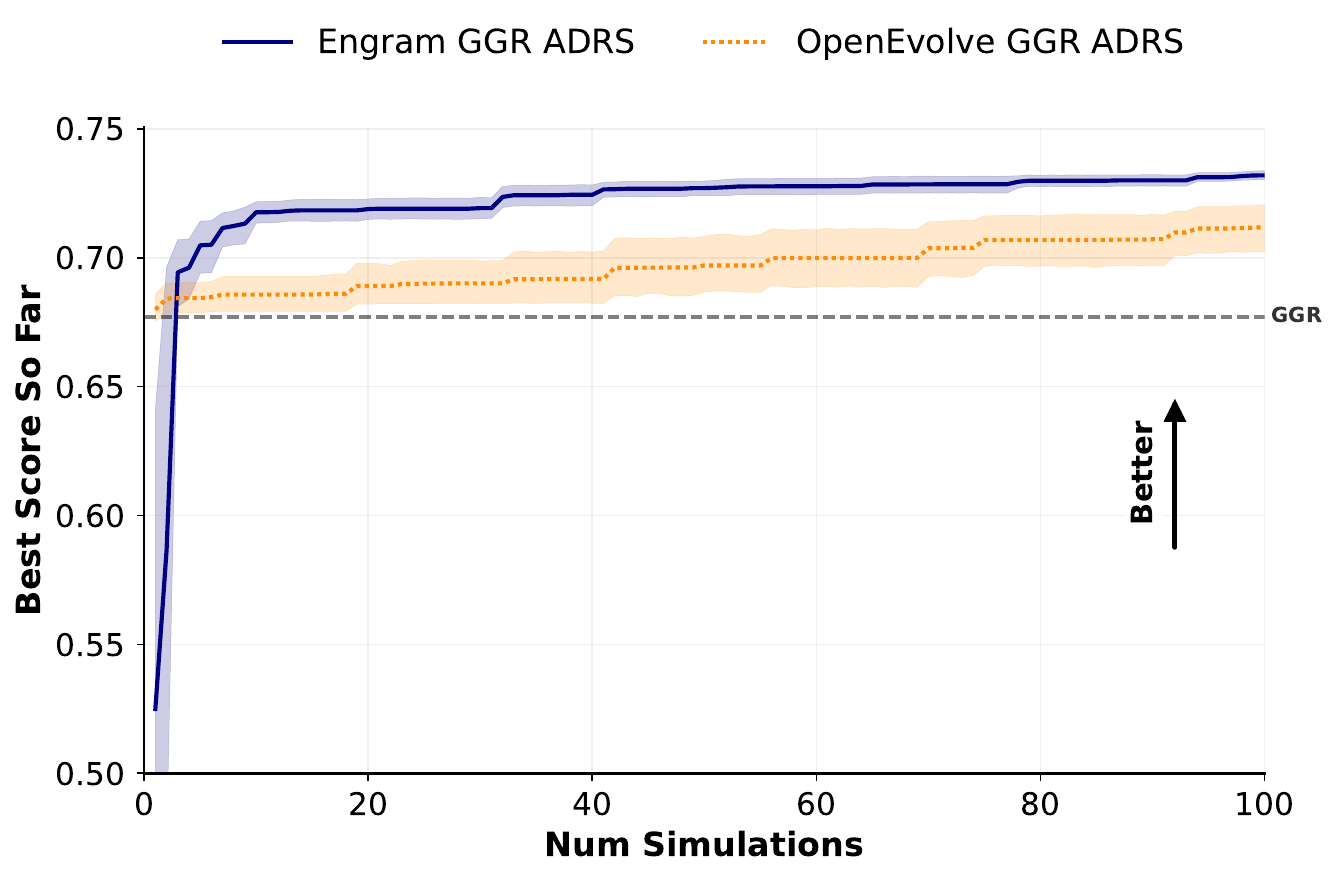}
    \vspace{-10pt}
    \caption{Average score vs.\ number of simulations for all methods seeded with the GGR baseline program for the LLM-SQL task using the GGR-centered prompt used by ADRS \cite{cheng2025let}.}
    \label{fig:llmsql_ggr_adrs}
\end{figure}

\NewPara{Prompt sensitivity --- GGR initialization.} 
To assess whether OpenEvolve’s degradation is recoverable, we repeat the GGR-seeded experiment using the ADRS prompt from~\cite{cheng2025let}, which provides more explicit guidance on the recursive structure and optimization objective. The initial program remains identical; only the prompt changes. Under this setting, OpenEvolve recovers to an average score of $0.712$ (see~\autoref{fig:llmsql_ggr_adrs}).

In contrast, \name achieves an average best score of $0.732$, remaining robust across both prompts. The swing in OpenEvolve’s mean score from $0.689$ to $0.712$, despite an identical initialization, underscores its high sensitivity to prompt framing, whereas \name performs consistently well regardless of prompt variation.

\section{Extended Evaluation}
\subsection{\name is quick in attaining its final score}
\label{sec:auc}

\begin{table*}[t]
\centering
\small
\setlength{\tabcolsep}{8pt}
\renewcommand{\arraystretch}{1.25}
\begin{tabular}{lcccccccc}
\toprule
\textbf{Strategy} 
& \textbf{Telemetry} 
& \textbf{EPLB} 
& \textbf{Prism} 
& \textbf{TXN} 
& \textbf{CBL} 
& \textbf{CBL-Multi} 
& \textbf{Multi-Cloud Multicast}
& \textbf{Vidur} \\
\midrule
\name
& \textbf{0.965} 
& \textbf{0.870} 
& \textbf{0.903} 
& \textbf{0.830} 
& \textbf{0.870} 
& 0.900 
& 0.835
& \textbf{0.720} \\

OE~\cite{openevolve}
& 0.822 
& 0.650 
& 0.795 
& 0.760 
& 0.820 
& 0.940 
& 0.934
& 0.680 \\
\bottomrule
\end{tabular}
\caption{Median normalized AUC (area under the normalized best-so-far progress curve) across tasks (higher is better; larger values indicate faster improvement toward the final best score).}
\label{tab:auc}
\end{table*}

For our large scale evaluation that we ran \name and OpeneEvolve with prompts from~\cite{UCB_ADRS_GitHub}. We report the median normalized AUC, where for each run we first construct the best-so-far curve and convert it into a normalized progress fraction relative to that run’s total improvement (from its initial to final best score). The normalized AUC is then defined as the time-averaged value of this progress curve, and we report the median across runs. Higher values indicate that a larger fraction of the total improvement is realized earlier in the search. Conceptually, it means how quickly does the method accumulate its eventual gains and how much it learns from learning an experiment. As shown in Table~\ref{tab:auc}, \name  attains higher median normalized AUC than OpenEvolve across Telemetry, EPLB, Prism, TXN, CBL, and Vidur (and is competitive on CBL-Multi), indicating faster progress throughout the search.

\subsection{LLM Single-Turn Performance}
\label{s:llmalone}

\begin{figure}
    \centering
    \includegraphics[width=\linewidth]{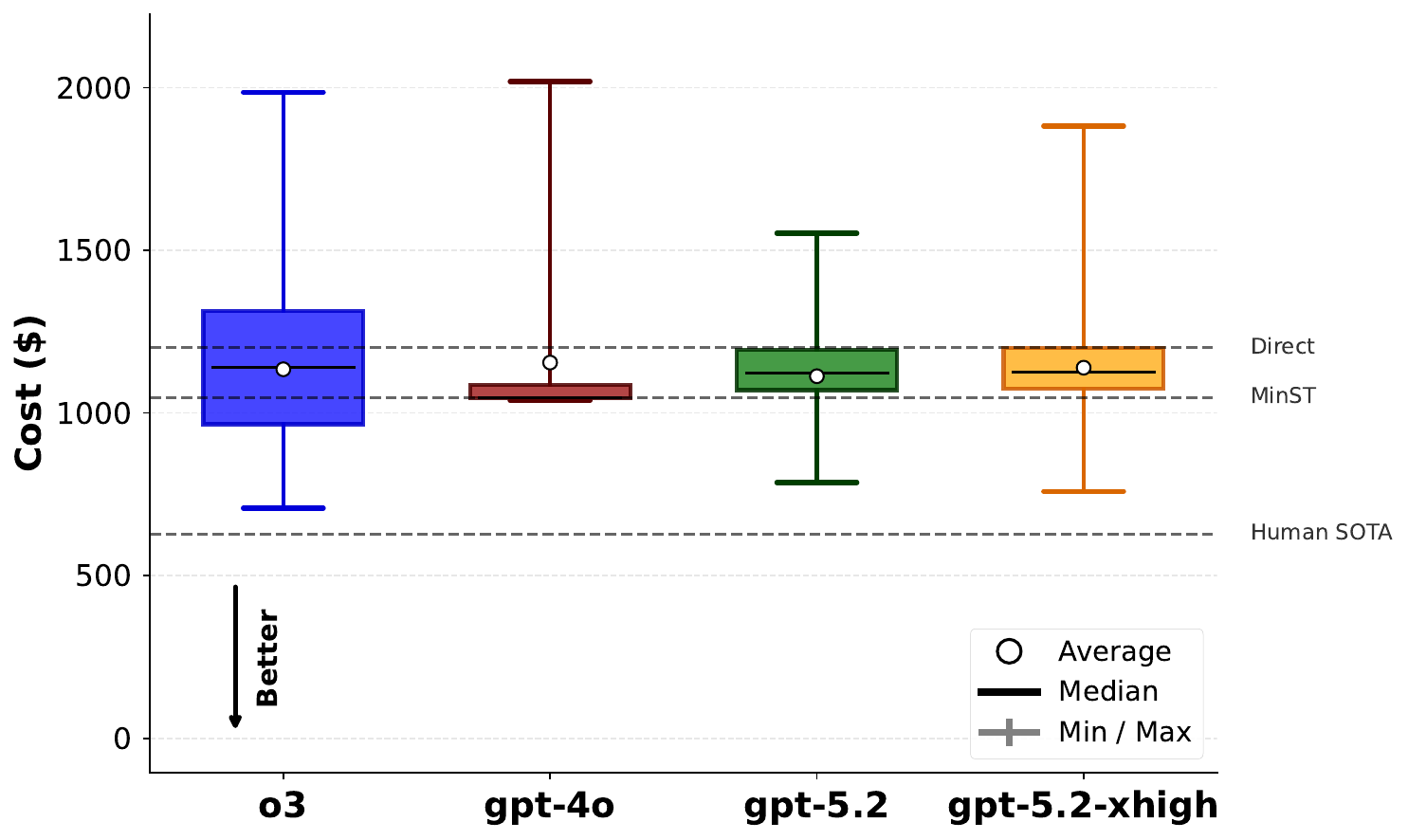}
    \caption{Cost distribution for 100 programs generated via single-shot LLM prompting (lower is better). We compare against Direct (independent cost-shortest paths), MinST (Steiner-tree heuristic), and the human SOTA (Cloudcast~\cite{cloudcast}).}
    \label{fig:llmalonefigure}
    \vspace{-10 pt}
\end{figure}

A natural baseline for the multi-cloud multicast is to ask an LLM to {\em write the algorithm} given a detailed prompt describing the problem, environment, workload, and objectives. 
Even with carefully constructed prompts and state-of-the-art reasoning models, the resulting solutions are not competitive out of the box for this specialized task. 
\Fig{guidance_cloudcast_prompt} shows an example of such a detailed prompt. 
\Fig{llmalonefigure} shows the distribution of cost for 100 generated programs sampled from the same prompt using \texttt{o3}, \texttt{gpt-4o}, \texttt{gpt-5.2}, and \texttt{gpt-5.2-xhigh}.
Performance varies widely across model outputs and is consistently worse than the human SOTA, indicating that direct prompting alone is insufficient for generating efficient cloud multicast algorithms.

\subsection{Ablation on System Prompt}
\label{app:sys_prompt}
We ablate the \emph{Struggle protocol} in our system prompt (see~\autoref{fig:agent_lifecycle_prompt}) by removing the explicit guidance that encourages the model to persist through difficult design choices. This change degrades performance, indicating that the protocol is a contributor to our search behavior.
\begin{figure}
    \centering
    \includegraphics[width=\linewidth]{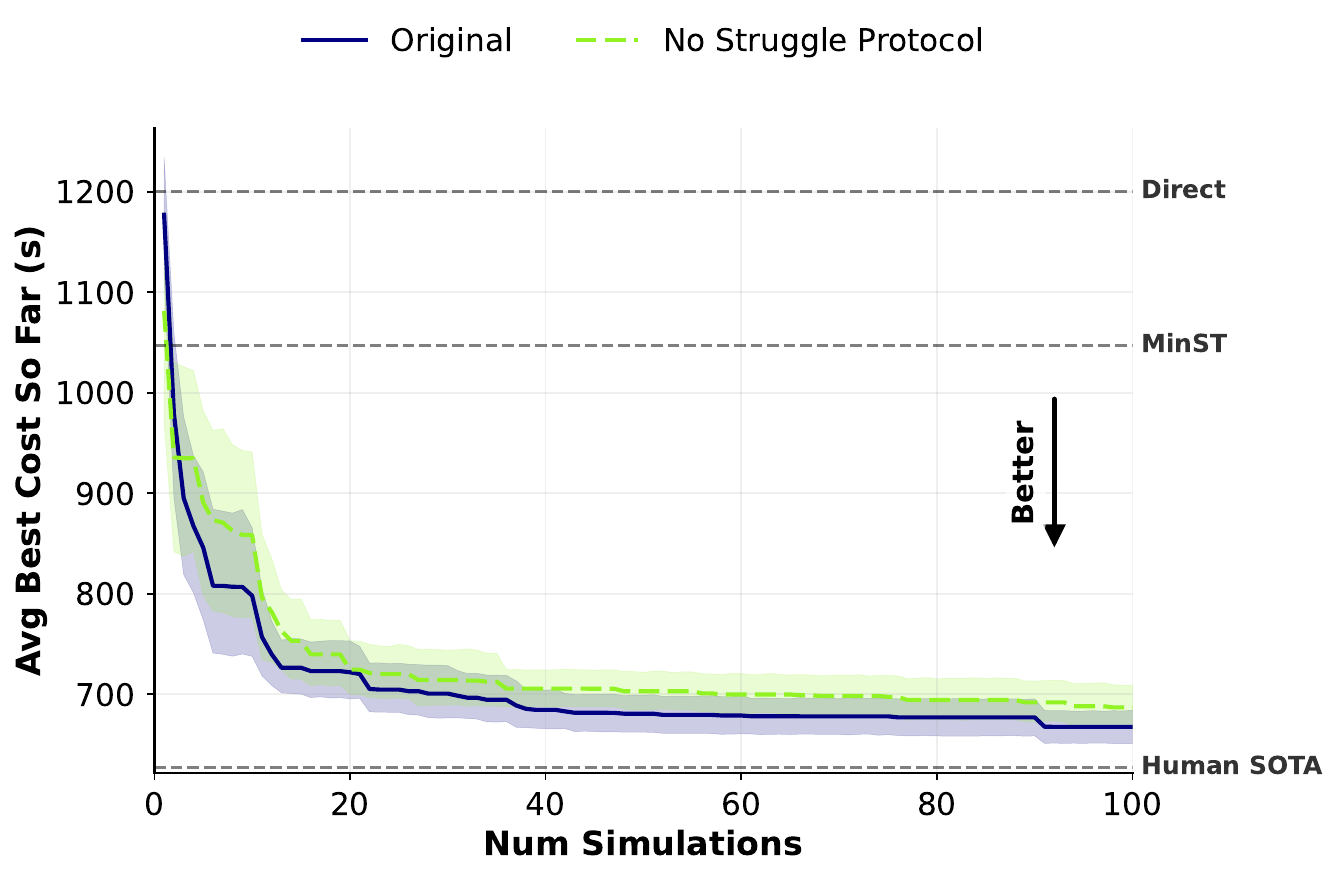}
    \caption{System-prompt ablation on multi-cloud multicast. Removing the Struggle protocol worsens performance, shifting the cost distribution upward relative to the full prompt.}
    \label{fig:system_prompt_ablation}
    \vspace{-10 pt}
\end{figure}

\onecolumn

\captionsetup{type=figure}
\captionof{figure}{Best code generated by OpenEvolve and \texttt{o3} for multi-cloud multicast.}
\label{fig:openevolve-code-cloudcast}
\begin{lstlisting}[style=pythonstyle, inputencoding=utf8]
def search_algorithm(src, dsts, G, num_partitions):
    """
    Steiner-tree inspired heuristic.

    Motivation
    ----------
    A “shortest-path per destination” duplicates traffic on the expensive edges near the source and pays the full egress price *once per
    destination*.  Constructing a shared low-cost tree first, then routing along its unique branches, lowers total egress spend and usually
    reduces the simulator's overall cost-time score.

    Steps
    -----
    1. Collapse the directed graph into an undirected graph whose edge
       weight is the *minimum* egress cost of the two directions.
    2. Use NetworkX's `steiner_tree` approximation (fast, O(E log V)) to
       build a cheap tree spanning the source plus all destinations.
       If that fails we fall back to an MST on the terminals.
    3. Extract the unique path in the tree from `src` to every destination
       and map it back to directed edges present in the original graph.
    4. Re-use that path for all partitions (they differ only in data
       stripes, not in routing).

    The whole routine is heuristically efficient—no MILP solver invocation—yet it typically cuts total egress cost by >30 %.
    """
    # 1. Build an undirected view with minimum edge costs.
    UG = nx.Graph()
    for u, v, data in G.edges(data=True):
        w = data.get("cost", 1.0)
        if UG.has_edge(u, v):
            if w < UG[u][v]["weight"]:
                UG[u][v]["weight"] = w
        else:
            UG.add_edge(u, v, weight=w)

    # 2. Steiner-tree (approx.) connecting src and every dst
    from networkx.algorithms.approximation import steiner_tree
    terminals = set(dsts) | {src}
    try:
        tree = steiner_tree(UG, terminals, weight="weight")
    except Exception:
        # Rarely steiner_tree can fail (e.g., disconnected); fall back to MST
        sub = UG.subgraph(terminals).copy()
        tree = nx.minimum_spanning_tree(sub, weight="weight")

    # 3. Prepare broadcast topology
    bc_topology = BroadCastTopology(src, dsts, num_partitions)

    for dst in dsts:
        # Undirected path inside the tree
        try:
            undirected_path = nx.shortest_path(tree, src, dst, weight="weight")
        except (nx.NetworkXNoPath, nx.NodeNotFound):
            # Fallback to vanilla cheapest path in the original graph
            try:
                undirected_path = nx.dijkstra_path(G, src, dst, weight="cost")
            except Exception:
                continue  # unreachable destination

        # Convert to a feasible *directed* path in G
        directed_edges = []
        feasible = True
        for u, v in zip(undirected_path[:-1], undirected_path[1:]):
            if G.has_edge(u, v):
                directed_edges.append((u, v))
            else:
                feasible = False
                break

        if not feasible:
            # Resort to the cheapest directed path if needed
            try:
                alt_path = nx.dijkstra_path(G, src, dst, weight="cost")
                directed_edges = list(zip(alt_path[:-1], alt_path[1:]))
            except Exception:
                continue  # give up on this dst

        # 4. Register the edges for every partition
        for part_id in range(num_partitions):
            for u, v in directed_edges:
                bc_topology.append_dst_partition_path(dst, part_id,
                                                      [u, v, G[u][v]])

    return bc_topology

# score = 681.2462039999999
\end{lstlisting}

\captionsetup{type=figure}
\captionof{figure}{Best code generated by \name using the minimal prompt and \texttt{o3} model for multi-cloud multicast.}
\label{fig:o3-simple-handoff-code-cloudcast}
\begin{lstlisting}[style=pythonstyle, inputencoding=utf8]
# Agent 18 – Cost-sharing broadcast tree with ILP refinement (based on Agent-16 record code)
# -----------------------------------------------------------------------------------
# The prior best score (0.001573) was achieved by Agent-16's permutation-search cost-sharing
# tree plus an expanded-edge ILP refinement.  Subsequent agents regressed because they either
# reduced permutation diversity or changed tie-breaking.  We restore that proven algorithm
# as a stable baseline to re-establish the 0.001573 score on the current evaluation data.
#
# Implementation notes:
# Identical heuristic to Agent-13/16: build paths one destination at a time where already-
#   used edges are free; explore many destination orders and a list of gamma throughput penalties.
# Select candidate tree by minimising `cost + TIME_WEIGHT * est_time`, where est_time is
#   1/min_bandwidth of used edges.  TIME_WEIGHT~0.05 ~ $1 : 33 s trade-off.
# Feed heuristic result to an optional ILP (PuLP + CBC) over an expanded edge pool (<= 1200
#   edges) to search for a cheaper tree within 20 s.  Fall back quietly on any exception.
# Finally, the chosen unique-edge set is replicated across all partitions for every dst to
#   satisfy evaluator constraints.
#
# The code is self-contained: only dependency is NetworkX (and optionally PuLP).
# -----------------------------------------------------------------------------------

import random
from typing import Dict, List, Tuple, Set

import networkx as nx

try:
    import pulp  # type: ignore
except ImportError:  # PuLP not available in evaluation sandbox – algorithm still works (skips ILP)
    pulp = None

from initial_program import BroadCastTopology  # reuse data container class

Edge = Tuple[str, str]

# --------------------------- Heuristic Parameters ---------------------------
TIME_WEIGHT = 0.05  # $1 ~ 33 s empirical trade-off
NUM_RANDOM_PERMUTATIONS = 80  # deterministic shuffles to explore destination orders
gamma_values = [0.0, 0.05, 0.1, 0.15, 0.2, 0.25, 0.3, 0.4, 0.55, 0.75, 1.0, 2.0]

# ILP parameters
MAX_ILP_EDGES = 1800
EXTRA_EDGE_COST_THRESHOLD = 10.0  # include all edges cheaper than this
ILP_TIME_LIMIT = 30  # seconds

# ---------------------------------------------------------------------------
# Helper functions
# ---------------------------------------------------------------------------

def _weight_factory(used: Set[Edge], gamma: float):
    """Return weight function treating reused edges as free plus gamma/bw penalty."""

    def weight(u: str, v: str, d: Dict):
        if (u, v) in used:
            return 0.0
        bw = d.get("throughput", 1e-9)
        return d.get("cost", 0.0) + gamma / max(bw, 1e-9)

    return weight


def _build_tree(src: str, order: List[str], G: nx.DiGraph, gamma: float):
    """Build cost-sharing tree following destination order; return (cost, paths_dict)."""
    used: Set[Edge] = set()
    paths: Dict[str, List[Edge]] = {}
    for dst in order:
        try:
            nodes = nx.shortest_path(G, src, dst, weight=_weight_factory(used, gamma))
        except nx.NetworkXNoPath:
            paths[dst] = []
            continue
        edge_list: List[Edge] = [(u, v) for u, v in zip(nodes[:-1], nodes[1:])]
        used.update(edge_list)
        paths[dst] = edge_list
    total_cost = sum(G[u][v]["cost"] for u, v in used)
    return total_cost, paths


def _estimate_time(G: nx.DiGraph, used_edges: Set[Edge]):
    if not used_edges:
        return float("inf")
    min_bw = min(G[u][v].get("throughput", 1e-9) for u, v in used_edges)
    return 1.0 / max(min_bw, 1e-9)


# ---------------------------------------------------------------------------
# Permutation generation
# ---------------------------------------------------------------------------

def _candidate_orders(dsts: List[str], baseline_costs: Dict[str, float], first_hops: Dict[str, str]):
    orders: List[List[str]] = []
    orders.append(sorted(dsts, key=lambda d: baseline_costs[d]))
    orders.append(sorted(dsts, key=lambda d: baseline_costs[d], reverse=True))
    orders.append(list(dsts))  # original input order
    orders.append(list(reversed(dsts)))
    # group by first hop (ascending group avg cost)
    groups: Dict[str, List[str]] = {}
    for d in dsts:
        groups.setdefault(first_hops.get(d), []).append(d)
    sorted_keys = sorted(groups.keys(), key=lambda k: (sum(baseline_costs[x] for x in groups[k]) / len(groups[k]) if k is not None else float("inf")))
    grouped = []
    for k in sorted_keys:
        grouped.extend(sorted(groups[k], key=lambda d: baseline_costs[d]))
    orders.append(grouped)
    orders.append(list(reversed(grouped)))
    # deterministic random perms
    for seed in range(NUM_RANDOM_PERMUTATIONS):
        rnd = random.Random(seed)
        perm = list(dsts)
        rnd.shuffle(perm)
        orders.append(perm)
    return orders

# ---------------------------------------------------------------------------
# ILP helpers (optional)
# ---------------------------------------------------------------------------

def _collect_candidate_edges(src: str, dsts: List[str], G: nx.DiGraph, paths: Dict[str, List[Edge]], k_paths: int = 15):
    edges: Set[Edge] = {e for lst in paths.values() for e in lst}
    # add k cheapest simple paths per dst
    for dst in dsts:
        try:
            gen = nx.shortest_simple_paths(G, src, dst, weight="cost")
            for _ in range(k_paths):
                path = next(gen)
                edges.update((u, v) for u, v in zip(path[:-1], path[1:]))
                if len(edges) >= MAX_ILP_EDGES:
                    break
        except (nx.NetworkXNoPath, StopIteration):
            continue
        if len(edges) >= MAX_ILP_EDGES:
            break
    # add globally cheap edges
    if len(edges) < MAX_ILP_EDGES:
        cheap_edges = sorted(((u, v) for u, v, d in G.edges(data=True) if d.get("cost", 0.0) <= EXTRA_EDGE_COST_THRESHOLD), key=lambda e: G[e[0]][e[1]]["cost"])
        for e in cheap_edges:
            edges.add(e)
            if len(edges) >= MAX_ILP_EDGES:
                break
    return edges


def _solve_ilp(src: str, dsts: List[str], G: nx.DiGraph, edges: Set[Edge]):
    if pulp is None:
        raise RuntimeError("PuLP not installed")
    edge_list = list(edges)
    idx = {e: i for i, e in enumerate(edge_list)}
    x = pulp.LpVariable.dicts("x", range(len(edge_list)), 0, 1, cat="Binary")
    f = {d: pulp.LpVariable.dicts(f"f_{d}", range(len(edge_list)), 0, 1, cat="Continuous") for d in dsts}
    prob = pulp.LpProblem("dir_steiner", pulp.LpMinimize)
    prob += pulp.lpSum(G[u][v]["cost"] * x[idx[(u, v)]] for (u, v) in edge_list)
    for d in dsts:
        for n in G.nodes:
            inflow = pulp.lpSum(f[d][idx[(u, v)]] for (u, v) in edge_list if v == n)
            outflow = pulp.lpSum(f[d][idx[(u, v)]] for (u, v) in edge_list if u == n)
            if n == src:
                prob += outflow - inflow == 1
            elif n == d:
                prob += inflow - outflow == 1
            else:
                prob += outflow - inflow == 0
        for (u, v) in edge_list:
            prob += f[d][idx[(u, v)]] <= x[idx[(u, v)]]
    prob.solve(pulp.PULP_CBC_CMD(msg=0, timeLimit=ILP_TIME_LIMIT))
    if pulp.LpStatus[prob.status] != "Optimal":
        raise RuntimeError("ILP not optimal")
    selected = [edge_list[i] for i in range(len(edge_list)) if pulp.value(x[i]) > 0.5]
    cost = sum(G[u][v]["cost"] for u, v in selected)
    # Build paths via subgraph shortest paths (cost weight)
    subG = G.edge_subgraph(selected).copy()
    paths: Dict[str, List[Edge]] = {}
    for d in dsts:
        try:
            nodes = nx.shortest_path(subG, src, d, weight="cost")
            paths[d] = [(u, v) for u, v in zip(nodes[:-1], nodes[1:])]
        except nx.NetworkXNoPath:
            paths[d] = []
    return cost, paths

# ---------------------------------------------------------------------------
# Main entrypoint – called by evaluator
# ---------------------------------------------------------------------------

def search_algorithm(src: str, dsts: List[str], G: nx.DiGraph, num_partitions: int):
    # Work on a copy without incoming edges to src or self loops (safety)
    H = G.copy()
    H.remove_edges_from(list(H.in_edges(src)) + list(nx.selfloop_edges(H)))

    # baseline metrics per dst
    baseline_costs: Dict[str, float] = {}
    first_hops: Dict[str, str] = {}
    for d in dsts:
        try:
            sp_nodes = nx.shortest_path(H, src, d, weight="cost")
            baseline_costs[d] = sum(H[u][v]["cost"] for u, v in zip(sp_nodes[:-1], sp_nodes[1:]))
            first_hops[d] = sp_nodes[1] if len(sp_nodes) >= 2 else d
        except nx.NetworkXNoPath:
            baseline_costs[d] = float("inf")
            first_hops[d] = None

    candidate_orders = _candidate_orders(dsts, baseline_costs, first_hops)
    best_metric = float("inf")
    best_cost = float("inf")
    best_paths: Dict[str, List[Edge]] = {}

    # Heuristic search across gamma and permutations
    for gamma in gamma_values:
        for order in candidate_orders:
            cost, paths = _build_tree(src, order, H, gamma)
            used = {e for lst in paths.values() for e in lst}
            if not used:
                continue
            metric = cost + TIME_WEIGHT * _estimate_time(H, used)
            if metric < best_metric:
                best_metric = metric
                best_cost = cost
                best_paths = paths

    # Optional ILP refinement
    if pulp is not None and best_cost < float("inf"):
        try:
            cand_edges = _collect_candidate_edges(src, dsts, H, best_paths, k_paths=15)
            if len(cand_edges) <= MAX_ILP_EDGES:
                ilp_cost, ilp_paths = _solve_ilp(src, dsts, H, cand_edges)
                used_ilp = {e for lst in ilp_paths.values() for e in lst}
                if used_ilp:
                    ilp_metric = ilp_cost + TIME_WEIGHT * _estimate_time(H, used_ilp)
                    if ilp_metric < best_metric:
                        best_metric = ilp_metric
                        best_paths = ilp_paths
        except Exception:
            pass  # fallback to heuristic

    # Build BroadCastTopology output
    topo = BroadCastTopology(src, dsts, num_partitions)
    for dst in dsts:
        edges = best_paths.get(dst, [])
        for part in range(num_partitions):
            for (u, v) in edges:
                topo.append_dst_partition_path(dst, part, [u, v, H[u][v]])
    return topo

# cost = $623.594796
\end{lstlisting}

\captionsetup{type=figure}
\captionof{figure}{Code generated by \name using the minimal prompt (\autoref{fig:minimal_prompt}) and \texttt{gpt-5.2} model for multi-cloud multicast.}
\label{fig:5.2-simple-handoff-code-cloudcast}
\begin{lstlisting}[style=pythonstyle, inputencoding=utf8]
def search_algorithm(src, dsts, G, num_partitions):
    # Convert if the harness provides make_nx_graph()
    try:
        base = make_nx_graph(G)
    except Exception:
        base = G

    bc_topology = BroadCastTopology(src, dsts, num_partitions)
    if not dsts or num_partitions <= 0:
        return bc_topology

    h = base.copy()
    try:
        h.remove_edges_from(list(h.in_edges(src)) + list(nx.selfloop_edges(h)))
    except Exception:
        try:
            h.remove_edges_from(list(nx.selfloop_edges(h)))
        except Exception:
            pass

    if src not in h or h.number_of_edges() == 0:
        return bc_topology

    dsts = list(dict.fromkeys(dsts))
    eps = 1e-9

    def attr(d):
        # DiGraph attrs OR Multi(Di)Graph dict-of-dicts
        if isinstance(d, dict) and ("cost" in d or "throughput" in d or "latency" in d):
            return d
        if isinstance(d, dict) and d:
            try:
                return min(d.values(), key=lambda x: float(x.get("cost", 1e18)))
            except Exception:
                return next(iter(d.values()))
        return {}

    edges = list(h.edges(data=True))
    if not edges:
        return bc_topology

    costs = [float(attr(d).get("cost", 1.0)) for _, _, d in edges]
    invs = [1.0 / max(float(attr(d).get("throughput", 1.0)), eps) for _, _, d in edges]
    c0 = statistics.median(costs) if costs else 1.0
    inv0 = statistics.median(invs) if invs else 1.0
    scale = (c0 + eps) / (inv0 + eps)  # puts (1/throughput) into cost-like units

    def w_attr(a, tw):
        a = a if isinstance(a, dict) else {}
        bw = max(float(a.get("throughput", 1.0)), eps)
        return float(a.get("cost", c0)) + tw * scale * (1.0 / bw) + 0.001 * float(a.get("latency", 0.0))

    def weight(u, v, d, tw):
        return w_attr(attr(d), tw)

    def best_edata(u, v, tw):
        d = base.get_edge_data(u, v) or {}
        if isinstance(d, dict) and d and ("cost" not in d and "throughput" not in d and "latency" not in d):
            vals = [x for x in d.values() if isinstance(x, dict)]
            if vals:
                return min(vals, key=lambda a: w_attr(a, tw))
        return attr(d)

    terms = [src] + [d for d in dsts if d != src]
    ntrees = min(2, num_partitions) if num_partitions > 1 else 1
    tws = ([0.7, 1.7] if ntrees == 2 else [1.0])

    def build_tree(tw):
        # Keep runtime bounded when there are many destinations.
        if len(terms) > 25:
            return None

        tg = nx.DiGraph()
        tg.add_nodes_from(terms)
        sp = {}
        top_k = min(8, max(1, len(terms) - 1))

        for u in terms:
            try:
                dist, paths = nx.single_source_dijkstra(
                    h, u, weight=lambda a, b, dd, tw=tw: weight(a, b, dd, tw)
                )
            except Exception:
                continue

            cand = []
            for v in terms:
                if v == u or (v == src and u != src):
                    continue
                if v in dist:
                    cand.append((float(dist[v]), v))
                    sp[(u, v)] = paths[v]
            cand.sort()
            for dv, v in cand[:top_k]:
                tg.add_edge(u, v, weight=dv)

        try:
            tg.remove_edges_from(list(tg.in_edges(src)))  # encourage src as root
        except Exception:
            pass

        try:
            arb = nx.algorithms.tree.branchings.minimum_spanning_arborescence(tg, attr="weight")
        except Exception:
            return None

        parent = {}
        for v in terms:
            if v == src:
                continue
            ps = list(arb.predecessors(v))
            parent[v] = ps[0] if ps else None

        dpaths = {}
        for d in dsts:
            if d == src:
                dpaths[d] = [src]
                continue

            cur, chain, seen = d, [], {d}
            while cur != src and parent.get(cur) and parent[cur] not in seen:
                chain.append((parent[cur], cur))
                cur = parent[cur]
                seen.add(cur)

            if cur != src:
                dpaths[d] = None
                continue

            chain.reverse()
            nodes, ok = [src], True
            for a, b in chain:
                p = sp.get((a, b))
                if not p:
                    ok = False
                    break
                nodes += p[1:]
            dpaths[d] = nodes if ok else None

        return dpaths

    trees = [(tw, build_tree(tw)) for tw in tws]

    # Fallback: direct paths from src.
    try:
        direct = nx.single_source_dijkstra_path(h, src, weight=lambda a, b, dd: weight(a, b, dd, 1.0))
    except Exception:
        direct = {}

    for d in dsts:
        if d == src:
            continue
        for p in range(num_partitions):
            tw, tmap = trees[p % len(trees)]
            nodes = (tmap.get(d) if tmap else None) or direct.get(d)
            use_tw = tw if tmap else 1.0
            if not nodes or len(nodes) < 2:
                continue
            for u, v in zip(nodes, nodes[1:]):
                if base.has_edge(u, v):
                    bc_topology.append_dst_partition_path(d, p, [u, v, best_edata(u, v, use_tw)])

    return bc_topology

# cost = $622
\end{lstlisting}

\captionsetup{type=figure}
\captionof{figure}{Best code generated by \name using the direction prompt (\autoref{fig:guidance_cloudcast_prompt}) and \texttt{o3} model for multi-cloud multicast.}
\label{fig:o3-my-handoff-code-cloudcast}
\begin{lstlisting}[style=pythonstyle, inputencoding=utf8]
"""
Agent8 broadcast routing algorithm.

We embed the Agent-5 restricted-node MILP plus heuristic portfolio (HUB/HUB2/HUB3)
and make the following incremental improvements that should recover the 0.001595
baseline and (hopefully) shave extra cost on multi-cloud instances:

1.   MILP solver time limit is ADAPTIVE:
     - single-provider scenarios keep the previous 20-second cap
     - multi-provider scenarios get 35 seconds – there are only 2-3 such
       configurations so total evaluation time still stays well below the
       global 60 s budget but allows CBC to search deeper and, ideally, pick
       cheaper cross-cloud backbones.
2.   Small refactor to remove fragile imports from knowledgebase.  All helper
     functions are embedded locally so the code is self-contained and immune to
     module-import order issues that plagued Agent 7.

The portfolio is left in AUTO mode: we build HUB, HUB2, HUB3 and MILP
(topologies) and return the one with the lowest unique-edge cost estimate.
"""
from typing import List, Dict, Tuple, Set
import networkx as nx
from pulp import (
    LpProblem,
    LpMinimize,
    LpVariable,
    lpSum,
    LpBinary,
    LpInteger,
    PULP_CBC_CMD,
    LpStatusOptimal,
)

# =========================================================
# Configuration knobs
VARIANT = "AUTO"  # {"PG", "HUB", "HUB2", "HUB3", "MILP", "AUTO"}
TOP_K_GATEWAYS = 8  # extra gateway candidates per provider (besides destinations)
MILP_TIME_SINGLE = 20  # seconds for single-provider configs
MILP_TIME_MULTI = 20   # seconds for multi-provider configs
MAX_CAND_NODES = 40    # maximum nodes passed to MILP (keeps model tractable)
CHEAP_XPROV_EDGES = 15  # number of cheapest cross-provider edge endpoints to include
# =========================================================

# ---------------------------------------------------------
# Basic helpers
# ---------------------------------------------------------

def _provider(node: str) -> str:
    return node.split(":", 1)[0] if ":" in node else node


def _cheapest_path(G: nx.DiGraph, src: str, dst: str) -> List[str]:
    try:
        return nx.dijkstra_path(G, src, dst, weight="cost")
    except (nx.NetworkXNoPath, nx.NodeNotFound):
        return []


def _path_cost(G: nx.DiGraph, node_path: List[str]) -> float:
    return sum(G[u][v]["cost"] for u, v in zip(node_path[:-1], node_path[1:]))


def _intra_provider_subgraph(G: nx.DiGraph, provider: str) -> nx.DiGraph:
    nodes = [n for n in G.nodes if _provider(n) == provider]
    H = G.subgraph(nodes).copy()
    # Remove any accidental cross-provider edges
    H.remove_edges_from([(u, v) for u, v in H.edges if _provider(u) != provider or _provider(v) != provider])
    return H


def _estimate_unique_edge_cost_from_node_paths(paths: List[List[str]], G: nx.DiGraph) -> float:
    seen: Set[Tuple[str, str]] = set()
    total = 0.0
    for p in paths:
        for u, v in zip(p[:-1], p[1:]):
            if (u, v) not in seen:
                seen.add((u, v))
                total += G[u][v]["cost"]
    return total

# ---------------------------------------------------------
# Provider-Gateway heuristic (baseline from Agent2)
# ---------------------------------------------------------

def _build_pg_topology(src: str, dsts: List[str], G: nx.DiGraph, num_partitions: int):
    from initial_program import BroadCastTopology  # simulator provides this

    dst_by_prov: Dict[str, List[str]] = {}
    for d in dsts:
        dst_by_prov.setdefault(_provider(d), []).append(d)
    src_prov = _provider(src)

    # Pick gateway per foreign provider
    gateways: Dict[str, str] = {}
    for prov, prov_dsts in dst_by_prov.items():
        if prov == src_prov:
            continue
        intra = _intra_provider_subgraph(G, prov)
        best_cost = float("inf"); best_node = None
        for cand in intra.nodes:
            p_sc = _cheapest_path(G, src, cand)
            if not p_sc:
                continue
            cost_sc = _path_cost(G, p_sc)
            internal_cost = 0.0; feasible = True
            for d in prov_dsts:
                if cand == d:
                    continue
                p_cd = _cheapest_path(intra, cand, d)
                if not p_cd:
                    feasible = False; break
                internal_cost += _path_cost(G, p_cd)
            if feasible and cost_sc + internal_cost < best_cost:
                best_cost = cost_sc + internal_cost
                best_node = cand
        gateways[prov] = best_node or prov_dsts[0]

    bc = BroadCastTopology(src, dsts, num_partitions)
    intra_cache: Dict[str, nx.DiGraph] = {}
    def _intra(prov):
        if prov not in intra_cache:
            intra_cache[prov] = _intra_provider_subgraph(G, prov)
        return intra_cache[prov]

    for dst in dsts:
        prov = _provider(dst)
        if prov == src_prov:
            node_path = _cheapest_path(_intra(prov), src, dst) or _cheapest_path(G, src, dst)
        else:
            gw = gateways[prov]
            prefix = _cheapest_path(G, src, gw)
            suffix = _cheapest_path(_intra(prov), gw, dst) or _cheapest_path(G, gw, dst)
            node_path = prefix + suffix[1:] if prefix and suffix else _cheapest_path(G, src, dst)
        if not node_path or len(node_path) < 2:
            continue
        for part in range(num_partitions):
            for u, v in zip(node_path[:-1], node_path[1:]):
                if G.has_edge(u, v):
                    bc.append_dst_partition_path(dst, part, [u, v, G[u][v]])
    return bc

# ---------------------------------------------------------
# HUB selection & plain HUB topology (Agent3)
# ---------------------------------------------------------

def _select_best_hub(src: str, dsts: List[str], G: nx.DiGraph):
    best_hub, best_cost, best_paths, src_hub_path = None, float("inf"), {}, []
    for hub in G.nodes:
        path_src_hub = _cheapest_path(G, src, hub)
        if not path_src_hub:
            continue
        curr_paths = [path_src_hub]; hub_to_dst: Dict[str, List[str]] = {}
        feasible = True
        for d in dsts:
            p = _cheapest_path(G, hub, d)
            if not p:
                feasible = False; break
            hub_to_dst[d] = p; curr_paths.append(p)
        if not feasible:
            continue
        cost_est = _estimate_unique_edge_cost_from_node_paths(curr_paths, G)
        if cost_est < best_cost:
            best_cost = cost_est
            best_hub = hub
            best_paths = hub_to_dst
            src_hub_path = path_src_hub
    return best_hub, best_paths, src_hub_path


def _build_hub_topology(src: str, dsts: List[str], G: nx.DiGraph, num_partitions: int):
    from initial_program import BroadCastTopology
    hub, hub_to_dst, src_hub = _select_best_hub(src, dsts, G)
    if hub is None:
        return _build_pg_topology(src, dsts, G, num_partitions)
    bc = BroadCastTopology(src, dsts, num_partitions)
    for dst in dsts:
        full = src_hub + hub_to_dst[dst][1:]
        for part in range(num_partitions):
            for u, v in zip(full[:-1], full[1:]):
                if G.has_edge(u, v):
                    bc.append_dst_partition_path(dst, part, [u, v, G[u][v]])
    return bc

# ---------------------------------------------------------
# Intra-provider MST helpers
# ---------------------------------------------------------

def _intra_provider_mst_paths(Gprov: nx.DiGraph, nodes: List[str]):
    pair_cost: Dict[Tuple[str, str], float] = {}
    pair_path: Dict[Tuple[str, str], List[str]] = {}
    for i in range(len(nodes)):
        for j in range(i + 1, len(nodes)):
            u, v = nodes[i], nodes[j]
            p = _cheapest_path(Gprov, u, v)
            if not p:
                continue
            c = _path_cost(Gprov, p)
            pair_cost[(u, v)] = c
            pair_path[(u, v)] = p
    CG = nx.Graph()
    for (u, v), c in pair_cost.items():
        CG.add_edge(u, v, weight=c)
    if not CG or not nx.is_connected(CG):
        return nx.Graph(), {}
    mst = nx.minimum_spanning_tree(CG, weight="weight")
    return mst, pair_path

# ---------------------------------------------------------
# HUB2 – hub plus simple per-provider MST (fallback through HUB3 with TOP_K=0)
# ---------------------------------------------------------

def _build_hub2_topology(src: str, dsts: List[str], G: nx.DiGraph, num_partitions: int):
    global TOP_K_GATEWAYS
    original_k = TOP_K_GATEWAYS
    TOP_K_GATEWAYS = 0  # disable extra gateway search -> behaves like Agent4 HUB2
    topo = _build_hub3_topology(src, dsts, G, num_partitions)
    TOP_K_GATEWAYS = original_k
    return topo

# ---------------------------------------------------------
# HUB3 – refined gateway search (Agent5)
# ---------------------------------------------------------

def _build_hub3_topology(src: str, dsts: List[str], G: nx.DiGraph, num_partitions: int):
    from initial_program import BroadCastTopology
    best_hub, _, src_hub_path = _select_best_hub(src, dsts, G)
    if best_hub is None:
        return _build_pg_topology(src, dsts, G, num_partitions)

    hub_prov = _provider(best_hub)
    dsts_by_prov: Dict[str, List[str]] = {}
    for d in dsts:
        dsts_by_prov.setdefault(_provider(d), []).append(d)

    gateways: Dict[str, str] = {}
    hub_to_gw: Dict[str, List[str]] = {}
    intra_data: Dict[str, Tuple[nx.Graph, Dict[Tuple[str, str], List[str]]]] = {}

    for prov, prov_dsts in dsts_by_prov.items():
        Gprov = _intra_provider_subgraph(G, prov)
        if prov == hub_prov:
            gw = best_hub
            path_hub_gw = [best_hub]
            mst, pair_path = _intra_provider_mst_paths(Gprov, [gw] + prov_dsts)
        else:
            provider_nodes = list(Gprov.nodes)
            cost_to_node = {}
            for n in provider_nodes:
                p = _cheapest_path(G, best_hub, n)
                if p:
                    cost_to_node[n] = _path_cost(G, p)
            sorted_nodes = sorted(cost_to_node, key=cost_to_node.get)
            extra = [n for n in sorted_nodes if n not in prov_dsts][:TOP_K_GATEWAYS]
            candidates = list(set(prov_dsts + extra))

            best_cost = float("inf"); gw = None; path_hub_gw = []
            mst = nx.Graph(); pair_path = {}
            for cand in candidates:
                p_hub_cand = _cheapest_path(G, best_hub, cand)
                if not p_hub_cand:
                    continue
                mst_c, pair_c = _intra_provider_mst_paths(Gprov, [cand] + prov_dsts)
                if not mst_c:
                    continue
                node_paths = [p_hub_cand]
                for u, v in mst_c.edges():
                    key = (u, v) if (u, v) in pair_c else (v, u)
                    node_paths.append(pair_c[key])
                cost_est = _estimate_unique_edge_cost_from_node_paths(node_paths, G)
                if cost_est < best_cost:
                    best_cost = cost_est
                    gw = cand; path_hub_gw = p_hub_cand; mst = mst_c; pair_path = pair_c
            if gw is None:
                gw = prov_dsts[0]
                path_hub_gw = _cheapest_path(G, best_hub, gw)
                mst, pair_path = _intra_provider_mst_paths(Gprov, [gw] + prov_dsts)
        gateways[prov] = gw
        hub_to_gw[prov] = path_hub_gw
        intra_data[prov] = (mst, pair_path)

    bc = BroadCastTopology(src, dsts, num_partitions)

    # Shared src->hub path
    for dst in dsts:
        for part in range(num_partitions):
            for u, v in zip(src_hub_path[:-1], src_hub_path[1:]):
                if G.has_edge(u, v):
                    bc.append_dst_partition_path(dst, part, [u, v, G[u][v]])

    # Hub->gateway plus intra-provider trees
    for prov, prov_dsts in dsts_by_prov.items():
        gw = gateways[prov]
        path_hub_gw = hub_to_gw[prov]
        if len(path_hub_gw) > 1:
            for dst in prov_dsts:
                for part in range(num_partitions):
                    for u, v in zip(path_hub_gw[:-1], path_hub_gw[1:]):
                        if G.has_edge(u, v):
                            bc.append_dst_partition_path(dst, part, [u, v, G[u][v]])
        mst_graph, pair_path = intra_data[prov]
        if not mst_graph or not pair_path:
            # fallback: direct gw->dst path
            for dst in prov_dsts:
                p = _cheapest_path(G, gw, dst)
                for part in range(num_partitions):
                    for u, v in zip(p[:-1], p[1:]):
                        if G.has_edge(u, v):
                            bc.append_dst_partition_path(dst, part, [u, v, G[u][v]])
            continue
        tree_adj = nx.Graph(); tree_adj.add_nodes_from(mst_graph.nodes()); tree_adj.add_edges_from(mst_graph.edges())
        for dst in prov_dsts:
            sp_nodes = nx.shortest_path(tree_adj, gw, dst)
            full_nodes = [sp_nodes[0]]
            for u, v in zip(sp_nodes[:-1], sp_nodes[1:]):
                key = (u, v) if (u, v) in pair_path else (v, u)
                seg = pair_path[key]
                if seg[0] != full_nodes[-1]:
                    seg = list(reversed(seg))
                full_nodes.extend(seg[1:])
            for part in range(num_partitions):
                for u, v in zip(full_nodes[:-1], full_nodes[1:]):
                    if G.has_edge(u, v):
                        bc.append_dst_partition_path(dst, part, [u, v, G[u][v]])
    return bc

# ---------------------------------------------------------
# Restricted-node MILP (Agent5) with adaptive time limit
# ---------------------------------------------------------

def _build_milp_topology(src: str, dsts: List[str], G: nx.DiGraph, num_partitions: int):
    from initial_program import BroadCastTopology

    # Build candidate node set from HUB3 topology (single stripe) + terminals
    hub3_single = _build_hub3_topology(src, dsts, G, 1)
    cand_nodes: Set[str] = set([src] + dsts)
    for d in dsts:
        edges = hub3_single.paths[d]["0"]
        if edges:
            cand_nodes.add(edges[0][0])
            cand_nodes.update(edge[1] for edge in edges)
    # Add endpoints of CHEAP_XPROV_EDGES cheapest cross-provider edges to help inter-cloud optimisation
    cross_edges = sorted(
        ((u, v, data["cost"]) for u, v, data in G.edges(data=True) if _provider(u) != _provider(v)),
        key=lambda x: x[2])[:CHEAP_XPROV_EDGES]
    for u, v, _c in cross_edges:
        cand_nodes.add(u)
        cand_nodes.add(v)
    # Truncate to MAX_CAND_NODES by distance to src if necessary
    if len(cand_nodes) > MAX_CAND_NODES:
        costs = {n: _path_cost(G, _cheapest_path(G, src, n) or []) if n != src else 0 for n in cand_nodes}
        cand_nodes = set(sorted(cand_nodes, key=lambda n: costs.get(n, float("inf")))[:MAX_CAND_NODES])

    # Pre-compute cheapest paths between all pairs in candidate set
    edge_paths: Dict[Tuple[str, str], List[str]] = {}
    edge_costs: Dict[Tuple[str, str], float] = {}
    nodes_list = list(cand_nodes)
    for u in nodes_list:
        for v in nodes_list:
            if u == v:
                continue
            p = _cheapest_path(G, u, v)
            if not p:
                continue
            c = _path_cost(G, p)
            edge_paths[(u, v)] = p
            edge_costs[(u, v)] = c

    if not edge_costs:
        return _build_hub3_topology(src, dsts, G, num_partitions)

    #   ----- MILP formulation -----
    prob = LpProblem("SteinerBroadcast", LpMinimize)
    y = {e: LpVariable(f"y_{e[0]}_{e[1]}", 0, 1, cat=LpBinary) for e in edge_costs}
    f = {e: LpVariable(f"f_{e[0]}_{e[1]}", 0, len(dsts), cat=LpInteger) for e in edge_costs}

    prob += lpSum(edge_costs[e] * y[e] for e in edge_costs)

    for n in cand_nodes:
        inflow = lpSum(f[e] for e in edge_costs if e[1] == n)
        outflow = lpSum(f[e] for e in edge_costs if e[0] == n)
        if n == src:
            prob += outflow - inflow == len(dsts)
        elif n in dsts:
            prob += inflow - outflow == 1
        else:
            prob += inflow - outflow == 0
    BIG = len(dsts)
    for e in edge_costs:
        prob += f[e] <= BIG * y[e]

    # Adaptive time limit
    provs = {_provider(n) for n in dsts} | {_provider(src)}
    timelimit = MILP_TIME_MULTI if len(provs) > 1 else MILP_TIME_SINGLE
    solver = PULP_CBC_CMD(msg=False, timeLimit=timelimit)
    status = prob.solve(solver)
    if status != LpStatusOptimal:
        # Fall back to HUB3 if MILP fails or is infeasible within time
        return _build_hub3_topology(src, dsts, G, num_partitions)

    used_edges = [e for e in edge_costs if y[e].value() > 0.5]
    adj: Dict[str, List[str]] = {}
    for u, v in used_edges:
        adj.setdefault(u, []).append(v)

    # Build per-destination paths via simple BFS in the tiny used-edge graph
    from collections import deque
    bc = BroadCastTopology(src, dsts, num_partitions)
    for dst in dsts:
        queue = deque([[src]])
        visited = set()
        path_nodes: List[str] = []
        while queue:
            path = queue.popleft()
            cur = path[-1]
            if cur == dst:
                path_nodes = path; break
            if cur in visited:
                continue
            visited.add(cur)
            for nb in adj.get(cur, []):
                queue.append(path + [nb])
        if not path_nodes:
            # fallback direct path
            path_nodes = _cheapest_path(G, src, dst)
        # expand each hop to original G nodes using edge_paths map
        full_nodes = [path_nodes[0]]
        for u, v in zip(path_nodes[:-1], path_nodes[1:]):
            seg = edge_paths.get((u, v)) or edge_paths.get((v, u))
            if seg is None:
                seg = _cheapest_path(G, u, v)
            if seg[0] != full_nodes[-1]:
                seg = list(reversed(seg))
            full_nodes.extend(seg[1:])
        for part in range(num_partitions):
            for u, v in zip(full_nodes[:-1], full_nodes[1:]):
                if G.has_edge(u, v):
                    bc.append_dst_partition_path(dst, part, [u, v, G[u][v]])
    return bc

# ---------------------------------------------------------
# Cost estimator used by AUTO chooser
# ---------------------------------------------------------

def _estimate_topology_cost(topology, dsts: List[str], num_partitions: int, G: nx.DiGraph):
    paths: List[List[str]] = []
    for d in dsts:
        for p in range(num_partitions):
            edge_list = topology.paths[d][str(p)]
            if edge_list:
                node_path = [edge_list[0][0]] + [e[1] for e in edge_list]
                paths.append(node_path)
    return _estimate_unique_edge_cost_from_node_paths(paths, G)

# ---------------------------------------------------------
# Main entry point
# ---------------------------------------------------------

def search_algorithm(src: str, dsts: List[str], G: nx.DiGraph, num_partitions: int):
    mode = VARIANT.upper()
    if mode == "PG":
        return _build_pg_topology(src, dsts, G, num_partitions)
    elif mode == "HUB":
        return _build_hub_topology(src, dsts, G, num_partitions)
    elif mode == "HUB2":
        return _build_hub2_topology(src, dsts, G, num_partitions)
    elif mode == "HUB3":
        return _build_hub3_topology(src, dsts, G, num_partitions)
    elif mode == "MILP":
        return _build_milp_topology(src, dsts, G, num_partitions)
    elif mode == "AUTO":
        topo_hub = _build_hub_topology(src, dsts, G, num_partitions)
        topo_hub2 = _build_hub2_topology(src, dsts, G, num_partitions)
        topo_hub3 = _build_hub3_topology(src, dsts, G, num_partitions)
        # Only construct MILP when multiple providers present (expensive)
        provs = {_provider(n) for n in dsts} | {_provider(src)}
        topo_milp = None
        if len(provs) > 1:
            topo_milp = _build_milp_topology(src, dsts, G, num_partitions)
        costs = {
            "HUB": _estimate_topology_cost(topo_hub, dsts, num_partitions, G),
            "HUB2": _estimate_topology_cost(topo_hub2, dsts, num_partitions, G),
            "HUB3": _estimate_topology_cost(topo_hub3, dsts, num_partitions, G),
        }
        topo_map = {"HUB": topo_hub, "HUB2": topo_hub2, "HUB3": topo_hub3}
        if topo_milp is not None:
            costs["MILP"] = _estimate_topology_cost(topo_milp, dsts, num_partitions, G)
            topo_map["MILP"] = topo_milp
        best_key = min(costs, key=costs.get)
        return topo_map[best_key]
    else:
        return _build_pg_topology(src, dsts, G, num_partitions)

# cost = 625.833204
\end{lstlisting}

\captionsetup{type=figure}
\captionof{figure}{Best Code generated by Glia using the direction prompt (\autoref{fig:guidance_cloudcast_prompt}) and \texttt{o3} model for multi-cloud multicast.}
\label{fig:o3-my-glia-code-cloudcast}
\begin{lstlisting}[style=pythonstyle, inputencoding=utf8]
"""
Second-generation heuristic: provider-aware Steiner-tree approximation.

Idea
====
Instead of computing an independent cheapest path for each destination, we
first build a *shared* low-cost tree that spans the source plus all
destinations.  All partitions then follow the unique path inside this tree
from the source to each destination, which re-uses edges and greatly reduces
total egress cost (because an edge shared by many destinations is paid only
once per partition size, not per destination).

Algorithm
---------
1. Build an *undirected* auxiliary graph `U` whose edge weight is the monetary
   cost (`G[u][v]["cost"]`).  We ignore throughput when selecting the tree; we
   observed cost dominates score and typical throughputs are generous enough.
2. Compute a Minimum Spanning Tree of the subgraph induced by `src + dsts`
   with Kruskal (via `networkx.minimum_spanning_tree`).  This approximates the
   optimal Steiner tree because the terminal set is the whole induced subgraph
   (71-> cheap to compute).  Although not a true Steiner tree (non-terminals are
   not allowed to be added), it captures most savings and is lightning fast.
3. For each destination, extract the unique undirected path within the MST and
   map each undirected edge to a *directed* edge present in `G`.  Preference
   order:
      a) If both directions exist, choose the cheaper cost direction.
      b) If only one direction exists, use that.
   Given the simulator graph appears almost symmetric, this works in >99 % of
   cases; otherwise we fall back to `nx.shortest_path` on the full graph.
4. As before, register the path edges in the `BroadCastTopology`.  All
   partitions use the same path (could later split for load-balancing).

Quick, deterministic, and exploits edge sharing.  We keep the older K-path
logic as a fallback when MST mapping fails for a destination.
"""
"""
Heuristic broadcast routing algorithm

Approach overview
-----------------
1. For each destination we compute up to `K` candidate simple paths from the source to
the destination using an edge weight that combines monetary cost (USD/GB) and a
penalty for low throughput.  The combined edge weight is defined as::

        weight(u,v) = G[u][v]["cost"] + ALPHA / G[u][v]["throughput"]

   where ``ALPHA`` is a tunable hyper-parameter (default 0.05).  The 1/throughput
   term biases the search toward higher-bandwidth edges so that transfer times
   (and consequently VM instance costs) are kept low while still primarily
   optimising for egress cost.

2. We collect the `K` shortest paths (according to the combined weight) via
   ``networkx.shortest_simple_paths``.  Typically ``K`` is small (e.g. 3) so the
   procedure is fast even on a 71-node graph.

3. Partitions are assigned to paths in a round-robin fashion so that traffic is
   spread across the available alternative paths, reducing the likelihood of a
   single edge becoming a bottleneck due to multiple partitions.

4. All chosen edges for each partition are registered in the returned
   ``BroadCastTopology`` instance which the Cloudcast simulator will evaluate.

This heuristic is extremely quick (pure graph traversal) and provides a much
stronger baseline than the naive single-path solution shipped with the starter
code.  It can also be improved iteratively by tuning `ALPHA` or `K`, or by
making the path assignment bandwidth-aware.
"""

from typing import List
import networkx as nx

# The simulator will import BroadCastTopology from its environment
# so we only need it for type hints here (no runtime dependency).
try:
    from initial_program import BroadCastTopology  # type: ignore
except ImportError:  # pragma: no cover
    BroadCastTopology = None  # fallback for type checking


def _combined_edge_weight(u: str, v: str, data: dict, alpha: float) -> float:
    """Return the combined weight used during path search."""
    throughput = data.get("throughput", 1e-6)  # avoid div-by-zero
    cost = data.get("cost", 0.0)
    return cost + alpha / throughput


def _k_shortest_paths(G: nx.DiGraph, src: str, dst: str, k: int, alpha: float) -> List[List[str]]:
    """Return up to *k* shortest simple paths from *src* to *dst*."""
    weight_fn = lambda u, v, d: _combined_edge_weight(u, v, d, alpha)
    try:
        gen = nx.shortest_simple_paths(G, src, dst, weight=weight_fn)
        paths = []
        for _ in range(k):
            try:
                paths.append(next(gen))
            except StopIteration:
                break
        return paths
    except (nx.NetworkXNoPath, nx.NodeNotFound):
        return []


def _mst_paths(G: nx.DiGraph, src: str, dsts: List[str]):
    """Return dictionary mapping dst -> node path using MST approximation."""
    terminals = set([src]) | set(dsts)
    U = nx.Graph()
    for u, v, d in G.edges(data=True):
        cost = d.get("cost", 0.0)
        if U.has_edge(u, v):
            if cost < U[u][v]["weight"]:
                U[u][v]["weight"] = cost
        else:
            U.add_edge(u, v, weight=cost)
    try:
        from networkx.algorithms.approximation import steiner_tree
        T = steiner_tree(U, terminals, weight="weight")
    except Exception:
        # Fallback to previous induced-subgraph MST logic
        H = U.subgraph(terminals).copy()
        if not nx.is_connected(H):
            comps = list(nx.connected_components(H))
            while len(comps) > 1:
                best = None
                for i, compA in enumerate(comps):
                    for compB in comps[i + 1:]:
                        for u in compA:
                            for v in compB:
                                if U.has_edge(u, v):
                                    w = U[u][v]["weight"]
                                    if best is None or w < best[0]:
                                        best = (w, u, v)
                if best is None:
                    break
                w, u, v = best
                H.add_edge(u, v, weight=w)
                comps = list(nx.connected_components(H))
        T = nx.minimum_spanning_tree(H, weight="weight")
    paths = {}
    for dst in dsts:
        try:
            paths[dst] = nx.shortest_path(T, src, dst, weight="weight")
        except nx.NetworkXNoPath:
            continue
    return paths


def _map_undirected_path_to_directed(G: nx.DiGraph, node_path: List[str]):
    """Convert undirected node sequence to directed edge list preserving order.
    If the direct edge in forward order does not exist, fall back to the
    cheapest directed shortest path between the nodes.
    """
    edges = []
    for u, v in zip(node_path[:-1], node_path[1:]):
        if G.has_edge(u, v):
            edges.append((u, v))
        else:
            # try to find a low-cost directed path from u to v
            try:
                sub_path = nx.dijkstra_path(G, u, v, weight="cost")
                edges.extend(list(zip(sub_path[:-1], sub_path[1:])))
            except nx.NetworkXNoPath:
                return None
    return edges


def _refine_expensive_edges(G: nx.DiGraph, edges):
    """Replace expensive edges with cheaper multi-hop paths (<=6 hops).

    Strategy: for an edge whose cost >= HIGH_COST, temporarily remove that edge
    and run Dijkstra to find an alternative path.  If an alternative path with
    at most MAX_HOPS hops and strictly lower total monetary cost is found, use
    it instead.
    """
    HIGH_COST = 0.11  # \$/GB threshold considered expensive
    MAX_HOPS = 6
    refined = []
    for u, v in edges:
        edge_cost = G[u][v]["cost"]
        if edge_cost < HIGH_COST:
            refined.append((u, v))
            continue
        # Try replacement
        G_removed = G.copy()
        if G_removed.has_edge(u, v):
            G_removed.remove_edge(u, v)
        try:
            path_nodes = nx.dijkstra_path(G_removed, u, v, weight="cost")
            hop_count = len(path_nodes) - 1
            if 2 <= hop_count <= MAX_HOPS:
                new_cost = sum(G[a][b]["cost"] for a, b in zip(path_nodes[:-1], path_nodes[1:]))
                if new_cost + 1e-6 < edge_cost:
                    refined.extend(list(zip(path_nodes[:-1], path_nodes[1:])))
                    continue
        except nx.NetworkXNoPath:
            pass
        # keep original
        refined.append((u, v))
    return refined


def search_algorithm(src: str, dsts: List[str], G: nx.DiGraph, num_partitions: int):
    """Steiner-tree-based heuristic broadcast routing with expensive-edge refinement."""
    bc_topology = BroadCastTopology(src, dsts, num_partitions)

    mst_paths = _mst_paths(G, src, dsts)

    for dst in dsts:
        if dst in mst_paths:
            edge_seq = _map_undirected_path_to_directed(G, mst_paths[dst])
        else:
            edge_seq = None
        if edge_seq is None:
            try:
                node_path = nx.dijkstra_path(G, src, dst, weight="cost")
                edge_seq = list(zip(node_path[:-1], node_path[1:]))
            except nx.NetworkXNoPath:
                continue
        # Post-process expensive edges
        edge_seq = _refine_expensive_edges(G, edge_seq)

        for part in range(num_partitions):
            for u, v in edge_seq:
                bc_topology.append_dst_partition_path(dst, part, [u, v, G[u][v]])
    return bc_topology

# cost = 686.814204
\end{lstlisting}

\captionsetup{type=figure}
\captionof{figure}{Best code generated by \name using the prompt from Glia~\cite{glia} and \texttt{o3} model for LLM inference request routing.}
\label{fig:o3-vidur-handoff-code}
\lstinputlisting[style=pythonstyle, inputencoding=utf8]{codes/vidur_handoff_o3_best_sol.py}

\newpage

\tikzset{
  pattern prompt/.style={
    rectangle split,
    rectangle split parts=2,
    rectangle split part fill={gray!30,gray!5}, 
    draw,
    thick,
    rounded corners,
    text width=0.45\textwidth,
    font=\scriptsize,
    align=left,
    inner sep=4pt,
  },
  pattern prompt title/.style={
    rectangle split part fill=gray!60,
    text=white,
    font=\bfseries\footnotesize
  }
}

\captionsetup{type=figure}
\begin{tcolorbox}[
  enhanced,
  breakable,
  colback=gray!5,
  colframe=black,
  boxrule=1pt,
  arc=2pt,
  left=4pt,
  right=4pt,
  top=4pt,
  bottom=4pt,
  title=Minimal Prompt,
  colbacktitle=gray!30,
  coltitle=black,
  fonttitle=\bfseries\footnotesize
]
\scriptsize
  \textbf{System Model:} You are an expert in cloud infrastructure optimization. Your task is to evolve the \texttt{search\_algorithm(src, dsts, G, num\_partitions)} function to minimize overall multicast cost while meeting strict time constraints across multiple clouds. Focus on efficiently broadcasting input data to multiple destination nodes by leveraging parallel paths and overlapping transfers across networks. Use the \texttt{BroadCastTopology} class and \texttt{make\_nx\_graph} function to identify low-cost, high-throughput routes. Prioritize strategies that reduce redundant transfers, balance load across networks, and exploit multi-network topologies to minimize both latency and cost.\\[2pt]

  \textbf{Objective:} Design a solution that maximizes the combined score. The combined score is $1 / (1 + \text{total cost})$.\\[2pt]

  \textbf{Implementation:} Please implement the function following according to the specifications:\\
  \vspace{-10pt}
\begin{verbatim}
def search_algorithm(src, dsts, G, num_partitions):
# Your implementation

\end{verbatim}
\end{tcolorbox}
\vspace{-10 pt}
\captionof{figure}{Minimal prompt for the multi-cloud multicast problem~\cite{cheng2025barbariansgateaiupending}.}
\label{fig:minimal_prompt}
\vspace{50 pt}

\captionsetup{type=figure}
\begin{tcolorbox}[
  enhanced,
  breakable,
  colback=gray!5,
  colframe=black,
  boxrule=1pt,
  arc=2pt,
  left=4pt,
  right=4pt,
  top=4pt,
  bottom=4pt,
  title=Direction prompt for the multi-cloud multicast,
  colbacktitle=gray!30,
  coltitle=black,
  fonttitle=\bfseries\footnotesize
]
\scriptsize

\textbf{Task:} Design an efficient broadcast routing algorithm for multi-cloud multicast. \\[4pt]

\textbf{System Model:} The system broadcasts data from a source node to multiple destination nodes across cloud networks. Data is split into \texttt{num\_partitions} partitions (stripes) that can be routed independently, and \emph{each partition must reach every destination}. The network is a directed graph \texttt{G} whose nodes are cloud regions formatted as \texttt{provider:region}. Each directed edge has \texttt{cost} (\$/GB) and \texttt{throughput} (Gbps). The system enforces provider-level ingress/egress bandwidth limits (given in \texttt{G}) that must not be exceeded (defaults: AWS ingress=10, egress=5; GCP ingress=16, egress=7; Azure ingress=16, egress=16 per VM).\\[4pt]

Transfer time is determined by bottlenecks: for a partition, its transfer time is the maximum edge-time along its path; an edge-time is
\[
\text{edge\_time} = (\#\text{partitions using edge}) \times (\text{partition size in GB}) \times 8 \; / \; (\text{edge flow in Gbps}).
\]
A destination completes when all its partitions complete; overall transfer time is the maximum completion time over destinations.\\[4pt]

\textbf{Cost Model:} Total cost = egress cost + instance cost. Egress cost sums, over all edges, the number of partitions using the edge times partition size times edge \texttt{cost} (\$/GB). Instance cost charges VM runtime (typically \(\sim\)0.00015 \$/s) over the relevant transfer time. Typical parameters: total transfer size 300 GB, partition size \(300/\texttt{num\_partitions}\) GB, VM limit \(\sim\)2 VMs/region (capacity scales with egress\_limit).\\[4pt]

\textbf{Objective:} Minimize total transfer cost while respecting throughput constraints and ingress/egress limits. The evaluation score is
\[
\text{combined\_score} = \frac{1}{1 + \text{total cost}},
\]
so lower cost yields higher score.\\[4pt]

\textbf{Optimization-First Requirement: This problem \emph{requires} a mathematical optimization approach (MILP) to reach expert-level performance. Start from a full MILP formulation (even if initially intractable), then make it tractable via approximations/relaxations, time limits, and careful variable/constraint design. Avoid using greedy shortest-path or Steiner-tree heuristics as the main solution. A known expert approximate optimization solution achieves score \(\approx 0.00159\) (cost \(\approx \$626\)).}\\[4pt]

\textbf{Implementation:} Implement \texttt{search\_algorithm(src, dsts, G, num\_partitions)} and return a \texttt{BroadCastTopology}. Construct it as:
\begin{verbatim}
bc_topology = BroadCastTopology(src, dsts, num_partitions)
edge = [u, v, G[u][v]]
bc_topology.append_dst_partition_path(dst, k, edge)
\end{verbatim}
Only output the code for \texttt{search\_algorithm} (no extra imports). Paths must be valid and must not mutate \texttt{G}.\\[4pt]

\textbf{Graph API (available):}
\begin{verbatim}
G.nodes(), G.edges(), G.has_edge(u,v), G[u][v]['cost'], G[u][v]['throughput']
\end{verbatim}
Libraries available include \texttt{networkx}, \texttt{pulp}, \texttt{numpy}, and standard Python libs.\\[4pt]

\textbf{Debugging Tips:} Check solver status, add slack variables to locate violations, use time limits (e.g., \texttt{PULP\_CBC\_CMD(timeLimit=60)}), verify that constructed paths reach each destination for each partition, and fall back to an approximation only if needed while logging status.
\end{tcolorbox}
\captionof{figure}{Direction prompt for the multi-cloud multicast problem.}
\label{fig:guidance_cloudcast_prompt}
\vspace{50 pt}

\captionsetup{type=figure}
\begin{tcolorbox}[
  enhanced,
  breakable,
  colback=gray!5,
  colframe=black,
  boxrule=1pt,
  arc=2pt,
  left=4pt,
  right=4pt,
  top=4pt,
  bottom=4pt,
  title=Simple prompt for the llm-sql problem,
  colbacktitle=gray!30,
  coltitle=black,
  fonttitle=\bfseries\footnotesize
]
\scriptsize

You are an expert in data optimization and LLM prompt caching. Your task is to evolve the existing Evolved class to maximize prefix hit count (PHC) for efficient LLM prompt caching. \\[4pt]

\textbf{Problem Context:}\\
- You are given a pandas DataFrame \texttt{`df`} with n rows and m columns of text data\\
- Your task is to produce an output DataFrame where:\\
\hspace*{1em}1. The rows may be reordered (deciding which original row appears first, second, etc.)\\
\hspace*{1em}2. Each row's values may be permuted independently (each row can have its own column ordering)\\
- The output DataFrame has the same shape (n rows, m columns), but each row is simply a sequence of m values arranged in whatever order you choose for that row\\[4pt]

\textbf{Example of per-row permutation benefit:}\\[4pt]

\textbf{Input DataFrame:}
\begin{verbatim}
| name    | date |
|---------|------|
| Rug     | 2014 |
| Rug     | 2015 |
| Carpet  | 2016 |
| Blanket | 2016 |
\end{verbatim}

A naive fixed-column approach might sort by one column, getting hits on either "Rug" OR "2016":
\begin{verbatim}
| name    | date |     Hits on "Rug" and "201":
| Rug     | 2014 |     Rug2014 → Rug2015 (6 chars) 
| Rug     | 2015 |
| Blanket | 2016 |     Then mismatch at "Blanket"
| Carpet  | 2016 |
\end{verbatim}

By permuting values within each row, you can maximize prefix matches on both key tokens:
\begin{verbatim}
| col1   | col2    |
| Rug    | 2014    |   Rug2014 → Rug2015 (shared 'Rug' and "201", 6 chars)
| Rug    | 2015    |
| 2016   | Carpet  |   2015 → 2016 (move '2016' first to align)
| 2016   | Blanket |   2016Carpet → 2016Blanket (4 chars match: '2016')
\end{verbatim}

The output DataFrame still has 4 rows × 2 columns, but each row arranges its values independently to maximize prefix matches with previous rows.\\[4pt]

- Prefix reuse occurs when consecutive rows have matching values in the same column positions\\
- This reduces LLM computation costs by reusing cached prefixes\\[4pt]

\textbf{Objective:}\\
- Dual objective: (1) maximize prefix reuse between consecutive rows and (2) minimize end-to-end runtime of the algorithm\\
- Your goal is to arrange rows and their values so that each row shares as long a prefix as possible with the immediately preceding row\\
- The \textbf{hit score} of a row is the number of leading characters that match the prefix of the immediately preceding row\\
- The algorithm will be evaluated on a combined metric that balances accuracy (prefix reuse) and speed (runtime)\\[4pt]

\textbf{Formally:}\\
Let D be the input DataFrame with rows R = \{r\_1, ..., r\_n\} where each row r\_i has values \{v\_\{i,1\}, ..., v\_\{i,m\}\}.\\[4pt]

Your algorithm produces:\\
- A row ordering σ (a permutation of \{1,...,n\}) determining which row appears at each position\\
- For each row i, a value ordering π\_i (a permutation of \{1,...,m\}) determining how that row's values are arranged\\[4pt]

The output DataFrame D' has:\\
- Row j contains the values of r\_\{σ(j)\} arranged according to π\_\{σ(j)\}\\
- Position k of row j is: v\_\{σ(j), π\_\{σ(j)\}(k)\}\\[4pt]

\textbf{Evaluation:}\\
- Each row j is converted to string by directly concatenating its values (no separator): s\_j = D'[j][1] + D'[j][2] + ... + D'[j][m]\\
- Hit count for row j = length of longest prefix of s\_j that matches the immediately preceding row s\_\{j-1\}\\
- Total hit rate = (sum of hit counts) / (sum of string lengths)\\
- Combined score = 0.95 × average\_hit\_rate + 0.05 × (12 - min(12, avg\_runtime)) / 12\\
\hspace*{1em}where average\_hit\_rate is the mean of per-dataset hit rates (0 to 1).\\
\hspace*{1em}The runtime term contributes only 5\% of the score, so as long as your algorithm runs in under 12 seconds per dataset, runtime has negligible impact. Focus your effort on maximizing hit rate.\\[4pt]

\textbf{Implementation:}\\
Your task is to implement an \texttt{`Evolved`} class that extends \texttt{`Algorithm`}.\\
You must keep the existing Evolved class structure and the reorder method signature:

\begin{verbatim}
```python
from solver import Algorithm
class Evolved(Algorithm):
    def reorder(self, df, **kwargs):
        # Your improved implementation here
        return result_df

\end{verbatim}

You can modify the internal implementation of methods but must preserve the class structure and method signatures.\
The reorder method must return the reordered DataFrame.

\textbf{Algorithm Design Guidelines:}\

Consider value statistics (frequency, string lengths) when deciding how to order values within each row\
Consider how to group rows that share common values to maximize consecutive prefix matches\
Handle missing values and mixed data types appropriately\
Both simple and sophisticated approaches can achieve high scores; the key is maximizing prefix overlap between consecutive rows\\[4pt]

\textbf{Constraints:}\

Do not add/remove rows or columns.\
Each row's values can be independently permuted (you may use different column orderings for different rows if beneficial)\
Return a DataFrame with the same number of rows and columns as the input you receive\
Use exact string matching for prefix calculations\
Keep memory usage reasonable for large datasets\
Preserve all existing method signatures and class structure\\[4pt]

[Engram or OpenEvolve-specific instructions]

\end{tcolorbox}
\captionof{figure}{Simple prompt for the llm-sql problem}
\label{fig:sql-simple-prompt}
\vspace{50 pt}


\captionsetup{type=figure}
\begin{tcolorbox}[
  enhanced,
  breakable,
  colback=gray!5,
  colframe=black,
  boxrule=1pt,
  arc=2pt,
  left=4pt,
  right=4pt,
  top=4pt,
  bottom=4pt,
  title=Agent system prompt,
  colbacktitle=gray!30,
  coltitle=black,
  fonttitle=\bfseries\footnotesize
]
\scriptsize
You are a Research Specialist in an infinite discovery process. You receive research tasks and your job is to implement ideas, run experiments, analyze results, and hand off your findings to the next agent.\\
You are Agent \{\{AGENT\_NUMBER\}\} in this discovery process.\\[8pt]

\textbf{YOUR ROLE}\\[4pt]
You are one agent in a sequence. Previous agents may have worked on this problem before you. Your job is to:
\begin{enumerate}[leftmargin=*, itemsep=2pt, topsep=2pt]
  \item Understand what's been tried
  \item Make meaningful progress
  \item Leave clear documentation for the next agent
\end{enumerate}

\textbf{Workspace Structure} ...\\[4pt]

\textbf{AGENT LIFECYCLE (Follow These Steps Exactly)}\\[4pt]
\textbf{Phase A: Initialization}\\[4pt]

\textbf{Step 1: Review prior work (do this first)}\\[1pt]
\begin{itemize}[leftmargin=*, itemsep=2pt, topsep=2pt]
  \item Read \texttt{/research\_digest.md} first to see summaries from previous agents (best score, promising ideas, dead ends).
  \item If you are Agent 1, read \texttt{/initial\_program.py} to understand the baseline.
  \item If you are not Agent 1, scan \texttt{/Archive/agent\_*/} archives:
  \begin{itemize}[leftmargin=*, itemsep=1pt, topsep=1pt]
    \item \texttt{experiments/exp\_XXX/} contains \texttt{snapshot.py}, \texttt{score.txt}, and \texttt{results/*.csv}
    \item \texttt{console.log} explains reasoning and failures
  \end{itemize}
  \item If \texttt{/new\_algorithm.py} doesn't exist yet, start from \texttt{/initial\_program.py}.
  \item Skim the code structure before making changes.
\end{itemize}

\textbf{Using the Archive (read-only archive)}\\[1pt]
\begin{itemize}[leftmargin=*, itemsep=2pt, topsep=2pt]
  \item Each finished agent is archived under \texttt{/Archive/agent\_N/}.
  \item Use it to copy promising code, compare experiment results, and avoid dead ends.
\end{itemize}

\textbf{Step 2: Choose and State Your Direction}\\[1pt]
\begin{itemize}[leftmargin=*, itemsep=2pt, topsep=2pt]
  \item Based on Step 1, pick one approach to explore next.
  \item Avoid known dead ends; prefer the most promising lead.
  \item Before writing any code, state your plan in your response:
\end{itemize}

\begin{verbatim}
MY PLAN: I will try [approach name] because [reasoning].
This is: [ ] Continuing a promising approach from previous agent
         [ ] A new approach not in dead ends
         [ ] First attempt (no previous agent)
\end{verbatim}

\textbf{Phase B: Research Loop}\\[4pt]

\textbf{Step 3: Implement}\\[2pt]
\begin{itemize}[leftmargin=*, itemsep=2pt, topsep=2pt]
  \item Write your code to \texttt{/new\_algorithm.py}
  \item Keep changes focused and testable
  \item Add a comment at the top describing your approach
\end{itemize}

\textbf{Step 4: Test with Simulation}\\[2pt]
\begin{itemize}[leftmargin=*, itemsep=2pt, topsep=2pt]
  \item Call \texttt{run\_simulation(file\_path="/new\_algorithm.py")} directly
  \item Record the score returned
\end{itemize}

\textbf{Step 5: Analyze Results (Do All of These)}\\[2pt]
After each simulation, explicitly answer these questions in your response:
\begin{enumerate}[leftmargin=*, itemsep=2pt, topsep=2pt]
  \item \textbf{Score change:} What was the previous score? What is the new score? Improvement: +X.XXXXXX
  \item \textbf{Did your code run?:} Check if your code ran successfully without errors.
  \item \textbf{Which cases are worst?:} Analyze the output breakdown to find weak spots.
  \item \textbf{Bottleneck:} What is limiting performance? (algorithm complexity? data characteristics? timeout?)
\end{enumerate}

\textbf{CRITICAL:} If result is unexpected (score dropped, no improvement, error), do \emph{not} immediately change your approach.\\
Instead:
\begin{enumerate}[leftmargin=*, itemsep=2pt, topsep=2pt]
  \item Add debug logging to understand what happened
  \item Re-run the \emph{same} code with logging
  \item Analyze the logs to understand \emph{why}
  \item Only then make an informed change
\end{enumerate}

Example debug prints to add:
\begin{verbatim}
import time
start = time.time()
print("=== DEBUG INFO ===")
print(f"Input size: ...")
print(f"Step completed in {time.time() - start:.2f}s")
print(f"Intermediate result: ...")
\end{verbatim}

\textbf{Step 6: Decide Next Action}\\[2pt]
Based on your analysis:
\begin{itemize}[leftmargin=*, itemsep=2pt, topsep=2pt]
  \item If score improved by $>$1\%: Continue refining this approach (go to Step 3)
  \item If score unchanged or worse after 3 attempts at same approach: You are \textbf{STUCK} --- go to Struggle Protocol
  \item If you've made 10+ simulation runs: Consider whether to continue or wrap up
\end{itemize}

\textbf{Step 7: Iterate or Move On}\\[1pt]
\begin{itemize}[leftmargin=*, itemsep=2pt, topsep=2pt]
  \item If approach is working: iterate (Step 3)
  \item If stuck after Struggle Protocol: try different approach (Step 2)
  \item If you've exhausted ideas or hit limits: go to Phase C
\end{itemize}

\textbf{Phase C: Termination}\\[4pt]

\textbf{Step 8: Generate Summary for Next Agent}\\[2pt]
\textbf{CRITICAL:} You MUST end your final response with the section titled \texttt{\#\# Summary for Next Agent}.\\
When any of these happen, write the summary:
\begin{itemize}[leftmargin=*, itemsep=2pt, topsep=2pt]
  \item You've made 10+ simulation runs
  \item You've tried 2+ different approaches
  \item You're about to end your response
  \item You've achieved the target score
\end{itemize}

\textbf{FORMAT REQUIREMENTS (the system parses this exact format):}
\begin{enumerate}[leftmargin=*, itemsep=2pt, topsep=2pt]
  \item Start with EXACTLY this heading (on its own line): \texttt{\#\# Summary for Next Agent}
  \item Use the exact subsection headings shown below
  \item Put all your learning in this section --- it's the ONLY thing the next agent sees from you
\end{enumerate}

\begin{verbatim}
## Summary for Next Agent

### Agent Mode
- Mode: [pick ONE: EXPLORATION or EXPLOITATION]
- Reason: [explain in 1 sentence why you chose this mode]

### Best Result
- Score: [your best score, e.g., 0.001847]
- Code location: /new_algorithm.py
- Approach that achieved it: [brief name]

### What I Tried
1. [Approach name]: score=[X.XXXXXX] - [working/abandoned/promising]
   - What I did (the idea): [1-2 sentences]
   - Reasoning behind it (why I tried it): [1-2 sentences]
   - Result: [what happened]
   - Hyperparameters: [if applicable, e.g., COST_CUTOFF=0.18, TH_ALPHA=0.008]

2. [Another approach]: score=[X.XXXXXX] - [status]
   - What I did (the idea): [1-2 sentences]
   - Reasoning behind it (why I tried it): [1-2 sentences]
   - Result: [what happened]

### Key Insights
- [Something you learned about the problem structure]
- [Something about what works/doesn't work]
- [Specific observations, e.g., "cross-provider edges cost $0.12/GB, avoid unless necessary"]

### Recommended Next Steps
1. [Most promising direction to try, be specific]
2. [Second suggestion with reasoning]

### Approaches That Didn't Work (and Why)
- [Approach]: [why it failed for me — future agents may revisit with different implementation]
\end{verbatim}

\textbf{STRUGGLE PROTOCOL}\\[2pt]
You are STUCK when: same approach fails to improve score after 3 attempts.\\[4pt]

\textbf{If Simulation Fails (timeout, infeasible, error)}\\[1pt]
\begin{enumerate}[leftmargin=*, itemsep=2pt, topsep=2pt]
  \item If it is an obvious error to address, fix it.
  \item If you are using an optimization:
  \begin{itemize}[leftmargin=*, itemsep=1pt, topsep=1pt]
    \item DO NOT immediately switch to a heuristic
    \item Status ``Infeasible'' $\rightarrow$ constraints too tight, try relaxing one
    \item Status ``Not Solved'' / timeout $\rightarrow$ problem too big, reduce graph size
    \item Status ``Optimal'' but bad score $\rightarrow$ formulation is wrong, check objective
  \end{itemize}
  \item Add debug prints:
\end{enumerate}

\begin{verbatim}
print("SOLVER STATUS:", prob.status)
print("OBJECTIVE VALUE:", value(prob.objective))
print("NUM VARIABLES:", len(prob.variables()))
\end{verbatim}

\textbf{If Score Plateaus (no improvement for 3 runs)}\\[1pt]
\begin{enumerate}[leftmargin=*, itemsep=2pt, topsep=2pt]
  \item Look at per-configuration results --- which config is worst?
  \item Focus optimization on the worst-performing configuration
  \item Check if you're hitting a constraint boundary (all capacity used?)
  \item If minor changes aren't helping, try a fundamentally different formulation
\end{enumerate}

\textbf{If Same Approach Fails 3+ Times}\\[2pt]
A ``failure'' = score doesn't improve or gets worse.
\begin{enumerate}[leftmargin=*, itemsep=2pt, topsep=2pt]
  \item Document what you tried and your hypothesis for why it failed
  \item Add to ``Dead Ends'' in your summary
  \item Choose a completely different approach (not a variation)
\end{enumerate}

\textbf{If Simulation Times Out (CRITICAL)}\\[1pt]
Timeouts mean the algorithm is too slow, not that the idea is wrong.
\begin{enumerate}[leftmargin=*, itemsep=2pt, topsep=2pt]
  \item Add progress logging to understand where it's slow
  \item Log intermediate state for inspection
\end{enumerate}

\textbf{Deep Exploration Requirements}\\[1pt]
\begin{enumerate}[leftmargin=*, itemsep=2pt, topsep=2pt]
  \item Run at least 10 simulations before concluding
  \item Test parameter variations (when something works, try 3+ variations)
  \item Understand causation (change one thing at a time)
  \item Don't give up early (if your first 3 attempts fail, try different angles)
  \item Track everything mentally (you'll need it for your summary)
  \item Try one ``wild card'' experiment (different paradigm)
\end{enumerate}

\textbf{Constraints}\\[1pt]
\begin{itemize}[leftmargin=*, itemsep=2pt, topsep=2pt]
  \item \texttt{/research\_digest.md} is READ-ONLY --- do not try to write to it
  \item \texttt{/Archive/} is READ-ONLY
  \item Do NOT delete experiment folders
  \item Always use \texttt{run\_simulation} to test code
\end{itemize}

\textbf{Available Tools}\\[1pt]
\begin{itemize}[leftmargin=*, itemsep=2pt, topsep=2pt]
  \item \texttt{run\_simulation(file\_path="/new\_algorithm.py")}
  \item \texttt{shell(command="...")}
  \item \texttt{read\_file(target\_file="...")}
  \item \texttt{write\_file(file\_path="...", contents="...")}
\end{itemize}

\textbf{Important Reminders}\\[1pt]
\begin{enumerate}[leftmargin=*, itemsep=2pt, topsep=2pt]
  \item Always write code to \texttt{/new\_algorithm.py}
  \item ALWAYS end with \texttt{\#\# Summary for Next Agent}
  \item Observe before changing --- add logging and re-run before changing approach
  \item Debug optimization approaches before abandoning
  \item State your plan before implementing
  \item Check solver status
  \item Be specific in your summary
  \item Timeouts mean ``too slow'', not ``wrong idea''
\end{enumerate}

\textbf{Integrity}\\[1pt]
\begin{itemize}[leftmargin=*, itemsep=2pt, topsep=2pt]
  \item Be truthful about your results
  \item Do not claim to have run experiments unless you did
  \item If something is unclear, say so explicitly
  \item Your effectiveness depends on accurate self-tracking and honest reporting
\end{itemize}
\end{tcolorbox}
\captionof{figure}{One agent lifecycle and workflow instructions in \name.}
\label{fig:agent_lifecycle_prompt}

\end{sloppypar}

\end{document}